\PassOptionsToPackage{dvipsnames}{xcolor}

\documentclass[11pt]{article}

\usepackage[final]{acl}

\usepackage[dvipsnames]{xcolor}
\usepackage{microtype}
\usepackage{inconsolata}
\usepackage[T1]{fontenc}
\usepackage[utf8]{inputenc}
\usepackage{latexsym}
\usepackage{times}

\usepackage{enumitem}
\setlist[itemize]{leftmargin=1.2em,itemsep=0em,topsep=0.3em}

\usepackage{placeins}
\usepackage{float}
\usepackage{graphicx}
\usepackage{svg}
\usepackage{wrapfig}
\usepackage{multicol}
\usepackage{multirow}
\usepackage{booktabs}

\usepackage{xurl}

\usepackage{soul}

\usepackage{usebib}
\newbibfield{editor}

\bibinput{custom}
\usepackage{float}
\usepackage{adjustbox}
\usepackage{changepage}
\usepackage{lipsum}
\usepackage{tikz}
\usetikzlibrary{shapes, backgrounds}
\usepackage{amsmath}
\usepackage{hyperref}
\usepackage{amsmath}
\usepackage{xfrac}
\usepackage{titlesec}
\usepackage{tcolorbox}
\usepackage{ifthen}
\usepackage{xspace}
\usepackage{etoolbox}
\usepackage{listings}

\titlespacing*{\section}{0pt}{\baselineskip}{0.5em}
\titlespacing*{\subsection}{0pt}{\baselineskip}{0.5em}
\titlespacing*{\paragraph}{0pt}{0.4em}{5pt}

\usepackage{caption}  
\newcommand{\anonymized}[1]{%
        #1%
}

\newcommand{\xmark}{%
\tikz[scale=0.23] {
    \draw[line width=0.7,line cap=round] (0,0) to [bend left=6] (1,1);
    \draw[line width=0.7,line cap=round] (0.2,0.95) to [bend right=3] (0.8,0.05);
}}
\newcommand{\cmark}{%
\tikz[scale=0.23] {
    \draw[line width=1.1,line cap=round] (0.25,0) to [bend left=10] (1,1);
    \draw[line width=1.2,line cap=round] (0,0.35) to [bend right=1] (0.23,0);
}}

\newcommand{\gray}[1]{\textcolor{gray}{#1}}

\newcommand{\coloredbox}[2]{
  \begin{adjustbox}{valign=c}
    \begin{tikzpicture}
      \node[
        fill=#1, 
        text=white, 
        rounded corners=0.3em,
        inner sep=0.02em,
        minimum height=0.92em,
        minimum width=0.92em,
        text centered, 
        anchor=center
      ] (n) {\textsf{\scriptsize{#2}}};
    \end{tikzpicture}
  \end{adjustbox}
}

\DeclareRobustCommand{\levelelem}{\coloredbox{Gray}{E}}

\DeclareRobustCommand{\levelhigh}{\coloredbox{ForestGreen}{H}}
\DeclareRobustCommand{\levelcoll}{\coloredbox{Cerulean}{C}}
\DeclareRobustCommand{\leveluni}{\coloredbox{Purple}{U}}
\DeclareRobustCommand{\levelolymp}{\coloredbox{Orange}{O}}

\newcommand{\examplebox}[2]{
    \begin{tcolorbox}[colframe=darkgray, colback=white, boxrule=0.5pt, width=0.9\linewidth,fontupper=\small,fontlower=\small\textit,title=#1.]
        #2
    \end{tcolorbox}\vspace{-1.2em}
}

\newcommand{\examplesplit}{
    \vspace{-0.1cm}%
    \tcblower%
    \vspace{-0.2cm}%
}

\newcommand{\autogrid}[1]{
    \begin{center}
    \def\arglist{#1} 
    \foreach \item [count=\i from 1] in \arglist {%
        \begin{minipage}[t]{0.49\textwidth}
            \vspace{0pt} 
            \item
        \end{minipage}
        \ifodd\i\hfill\else\\[0.4cm]\fi 
    }
    \end{center}
}

\makeatletter
\newcommand{\umath}{U-MATH\xspace}
\newcommand{\umatht}{U-MATH$_\text{T}$\xspace}
\newcommand{\umathtext}{U-MATH$_\text{Text}$\xspace}
\newcommand{\umathv}{U-MATH$_\text{V}$\xspace}
\newcommand{\umathvisual}{U-MATH$_\text{Visual}$\xspace}
\newcommand{\mumath}{$\boldsymbol{\mu}$-MATH\xspace}
\makeatother

\title{U-MATH: A University-Level Benchmark \\ for Evaluating Mathematical Skills in Large Language Models}

\author{Konstantin Chernyshev\thanks{Corresponding author: \texttt{kchernyshev@toloka.ai}},~Vitaliy Polshkov, Ekaterina Artemova, Sergei Tilga \\
    Toloka AI \\
    \texttt{\{kchernyshev, cogwheelhead, katya-art, tilgasergey\}@toloka.ai} \\
\AND
    Alex Myasnikov, Vlad Stepanov \\
    Gradarius \\
    \texttt{\{alex, vstepanov\}@gradarius.com} \\
\And
    Alexei Miasnikov \\
    Gradarius, Stevens Institute of Technology \\
    \texttt{amiasnik@stevens.edu} \\
}

\begin{document}

\maketitle

\begin{abstract}
    Current evaluations of mathematical skills in Large Language Models are constrained by benchmarks lacking scope, particularly for multi-modal problems --- frequently relying on school-level~\citep{cobbe2021training,lu2023mathvista,zhang2024mathverse}, niche Olympiad-style~\citep{fang2024mathodyssey,mao2024champ}, simple quiz format~\citep{yue2024mmmu,qiao2024we} or relatively small~\citep{lewkowycz2022solving} datasets.

    To address this, we introduce \textbf{U-MATH}, a novel benchmark comprising \textbf{1,100} unpublished open-ended university-level problems sourced from current US curricula, with \textbf{20\%} incorporating visual elements. Given the free-form nature of U-MATH problems, we employ LLM judges for solution evaluation and release \textbf{$\boldsymbol{\mu}$-MATH}, a meta-evaluation benchmark composed of \textbf{1,084} \umath-derived tasks enabling precise assessment of these judges.

    Benchmarking leading LLMs reveals marked limitations in multi-modal reasoning, with maximum accuracy reaching 93.1\% on textual tasks but only 58.5\% on visual ones. Furthermore, solution judgment proves challenging, requiring the most advanced models to achieve meaningfully high performance, even still peaking at an imperfect F1-score of 90.1\%.

         We open-source \umath, \mumath, and all our evaluation code.\footnote{\url{https://github.com/toloka/u-math}}
\end{abstract}

\begin{figure*}[!htbp]
    \centering
    \examplebox{Example: Differential Calculus}{%

        \textbf{\umath Problem:} \\
        The function $s(t) = 2 \cdot t^3 - 3 \cdot t^2 - 12 \cdot t + 8$ represents the position of a particle traveling along a horizontal line.

        1. Find the velocity and acceleration functions.\\
        2. Determine the time intervals when the object is slowing down or speeding up.

        \examplesplit
        {\footnotesize
        
        \textbf{Reference Solution (shortened):} \\
        The velocity is $v(t) = s'(t) = \boxed{6\cdot t^2 - 6\cdot t - 12}$, zeros of the $v(t)$ are $t = -1, 2$. \\
        The acceleration is $a(t) = v'(t) = \boxed{12\cdot t - 6}$, zero of the $a(t)$ is $t = \frac{1}{2}$.  

        It speeds up when $v(t)$ and $a(t)$ have the same sign, and slows down when opposite.

        $$
        \begin{array}{cccc}
        \hline
        \text{Interval} & v(t) & a(t) & \text{Behavior} \\
        \hline
        (-\infty, -1) & >0 & <0 & \text{Slowing down} \\
        (-1, \frac{1}{2}) & <0 & <0 & \text{Speeding up} \\
        (\frac{1}{2}, 2) & <0 & >0 & \text{Slowing down} \\
        (2, \infty) & >0 & >0 & \text{Speeding up} \\
        \hline
        \end{array}
        $$
            
        Accounting for non-negative time, speed up on $\boxed{(0, \sfrac{1}{2}) \text{ and } (2, \infty)}$, slow down on $\boxed{(\sfrac{1}{2}, 2)}$.
        }
        \vspace{-1em}
    }
    \caption{A \umath sample. A common students' error reported by the author is overlooking time non-negativity.}
    \vspace{-1em}
\end{figure*}

\section{Introduction} \label{sec:introductions}
Assessing the mathematical proficiency of Large Language Models (LLMs) is crucial for evaluating their fundamental reasoning capabilities \citep{ahn2024largelanguagemodelsmathematical}. The most widely used benchmarks, GSM8K \citep{cobbe2021training} and MATH \citep{hendrycks2021measuring}, primarily cover school-level problems, overlooking advanced topics and facing rapid saturation \citep{achiam2023gpt}. Although some MATH problems and other recent works introduce harder concepts, they are limited in size and scope, relying on competition-style problems and neglecting the practical middle-ground of university-level coursework.

There is also growing demand for visual reasoning assessment in multi-modal LLMs \citep{ahn2024largelanguagemodelsmathematical}. Datasets such as the recent MATH-V \citep{wang2024measuring} provide numerous visual problems but face similar topic limitations or rely on the multiple-choice format, making the tasks significantly easier \citep{li2024multiplechoicequestionsreallyuseful,pezeshkpour2023largelanguagemodelssensitivity}.

In turn, reliably evaluating complex free-form responses is challenging \citep{hendrycks2021measuring}, which results in LLM judges becoming the de facto standard despite known biases and inconsistencies \citep{zheng2023judging}. These biases are often overlooked and unquantified, preventing potential correction. Quantifying auto-evaluation errors requires datasets designed specifically to assess the evaluators themselves, also called meta-evaluations. While mathematical meta-evaluation datasets do exist, they are mostly based on GSM8K and MATH, inheriting their scope limitations.

\vspace{0.3em}
\noindent
To address these gaps, we introduce the \textbf{\umath} (\textit{U}niversity \textit{Math}) and $\pmb{\mu}$\textbf{-MATH} (\textit{M}eta \textit{U}-\textit{MATH}) benchmarks. Our main contributions are:

\begin{enumerate}
    \setlength\itemsep{0.2em}

    \item \textbf{\umath} (Section~\ref{sec:methodology}): We open-source 1,100 university-level problems, balanced across six core university subjects. The problems are collected from actual coursework and supplied with correct answers, with approximately 20\% incorporating visual elements.

    \item \textbf{\mumath} (Section~\ref{subsec:meta-eval}): We introduce a set of 1084 meta-evaluation tasks designed to assess the quality of LLM judges by selecting approximately 25\% of the \umath problems, supplying each with four solutions produced by four different top-performing language models, and providing ground truth labels on generated solutions' correctness.

    \item \textbf{Comparative analysis} (Section~\ref{sec:experiments}): We compare various open-source and proprietary LLMs on \umath and \mumath, revealing significant deficiencies in solving university-level multi-modal problems. We also find proprietary models to outperform open-source ones on these tasks, while near-parity is observed with the text modality. Judgment also proves challenging for LLMs, with only the best-performing and most recent models attaining adequately high scores. In addition, we demonstrate that most current systems exhibit biased and unstable judgment performance. Finally, we establish that judgment as a skill is distinct from problem-solving and identify characteristic behavioral tendencies in LLM judges.
\end{enumerate}

\noindent
We release the \umath and \mumath benchmarks under a permissive license to facilitate further research and ensure reproducibility.

\section{Background} \label{sec:related_work}

Evaluating mathematical capabilities of LLMs is an essential direction of AI research \citep{ahn2024largelanguagemodelsmathematical}. Apart from mathematical proficiency being important in and of itself, studies show that fine-tuning with math and code-related data enhances models' fundamental `cognitive skills' \citep{prakash2024finetuningenhancesexistingmechanisms} and reasoning capabilities \citep{chen2024controlmathcontrollabledatageneration}, further necessitating the creation of mathematical evaluation datasets. Despite significant progress, many existing datasets are limited in scope, complexity of the problems, or size, as evidenced by the summary in Table~\ref{table:existing_datasets}.

\begin{table*}[!ht]
    \centering
    \begin{adjustbox}{width=0.85\linewidth}
    \begin{tabular}{llllllll}
    \toprule
    \multirow{2}{*}{\textbf{Dataset}} & \multirow{2}{*}{\textbf{Levels}} & \multirow{2}{*}{\textbf{\%Uni. Level}} & \multirow{2}{*}{\textbf{\#Test}} & \multirow{2}{*}{\textbf{\%Visual}} & \multirow{2}{*}{\textbf{\%Free-form}} & \textbf{\#Free-form Text-only} & \textbf{\#Free-form Visual} \\
    & & & & & & \textbf{~~Uni. Level Test} & \textbf{~~Uni. Level Test} \\
    \midrule
    MMLU$_\textit{Math}$ \citep{hendrycks2020measuring} & \levelelem\levelhigh\levelcoll & 0 & 1.3k & 0 & 0 & 0 & 0\\
    GSM8k \citep{cobbe2021training} & \levelelem & 0 & 1k & 0 & 0 & 0 & 0 \\
    MATH \citep{hendrycks2021measuring} & \levelhigh\levelolymp & 0 & 5k & 0 & 100 & 0 & 0 \\
    MiniF2F \citep{zheng2021minif2f} & \levelelem\levelhigh\levelolymp & 0 & 244 & 0 & 100 & 0 & 0 \\
    OCWCourses \citep{lewkowycz2022solving} & \leveluni & 100 & 272 & 0 & 100 & 272 & 0 \\
    ProofNet \citep{azerbayev2023proofnet} & \levelcoll\leveluni & $\approx$50 & 371 & 0 & 100 & $\approx$180 & 0 \\
    CHAMP \citep{mao2024champ} & \levelhigh\levelolymp & 0 & 270 & 0 & 100 & 0 & 0 \\
    MathOdyssey \citep{fang2024mathodyssey} & \levelhigh\leveluni\levelolymp & $\approx$25 & 387 & 0 & 100 & $\approx$50 & 0 \\
    \midrule
    MMMU$_\textit{Math}$ \citep{yue2024mmmu} & \levelcoll & 0 & 505 & 100 & 0 & 0 & 0 \\
    MathVista \citep{lu2023mathvista} & \levelelem\levelhigh\levelcoll & 0 & 5k & 100 & 46 & 0 & 0 \\
    MATH-V \citep{wang2024measuring} & \levelelem\levelhigh\levelolymp & 0 & 3k & 100 & 50 & 0 & 0 \\
    We-Math \citep{qiao2024we} & \levelelem\levelhigh\leveluni & $\approx$20 & 1.7k & 100 & 0 & 0 & 0 \\
    MathVerse \citep{zhang2024mathverse} & \levelhigh & 0 & 4.7k & 83.3 & 45 & 0 & 0 \\
    \midrule
    \textbf{\umath} (this work) & \leveluni & 100 & 1.1k & 20 & 100 & \textbf{900} & \textbf{200} \\
    \bottomrule
    \end{tabular}
    \end{adjustbox}
    \caption{Existing auto-evaluated math benchmarks along with their sizes, visual sample percentages, and open-ended problem percentages. Level markers: \levelelem Elementary to Middle School, \levelhigh High School, \levelcoll College, \leveluni University, \levelolymp Olympiads.}
    \label{table:existing_datasets}
\end{table*}

\paragraph{Textual Mathematical Benchmarks.} Datasets like MathQA \citep{amini2019mathqa} and the mathematics subset of MMLU \citep{hendrycks2020measuring} represent early efforts to assess math capabilities of LLMs, relying primarily on rather simple multiple-choice problems. Today, even smaller models have achieved high scores with these tasks \citep{li2024common7blanguagemodels}, rendering the benchmarks obsolete.

Subsequently, more comprehensive datasets emerged, including GSM8K \citep{cobbe2021training}, MATH \citep{hendrycks2021measuring}, and MGSM \citep{shi2022language} (a multilingual version of 250 GSM8K samples). These, however, mostly include elementary- to high-shool level problems, which may not fully gauge the depth of mathematical reasoning, and quickly approach saturation as well.

Recent works aim to introduce more advanced concepts, prominent examples including MathOdyssey \citep{fang2024mathodyssey} and CHAMP \citep{mao2024champ}, composed primarily of problems from high-school competitions, ProofNet \citep{azerbayev2023proofnet} and MiniF2F \citep{zheng2021minif2f}, focused on formal proof composition and auto-formalization, and OCWCourses \citep{lewkowycz2022solving}, based on MIT curricula contents. However, these datasets are constrained by their smaller sizes (under 400 problems each), and most focus on Olympiad-style problems, missing the more practical topics of university coursework. Apart from that, all of them rely on publicly available materials, allowing for data leakage.

\noindent
Our dataset offers \textbf{over three times more open-ended university-level problems} compared to these existing alternatives, with all of its problems previously unpublished.




\paragraph{Visual Mathematical Benchmarks.} With the rise of multi-modal LLMs, demand for visual mathematical benchmarks is growing \citep{zhang2024mathverse, qiao2024we}. Early efforts focused primarily on simpler geometry problems, as seen with datasets such as GeoQA \citep{chen2022geoqa}, UniGeo \citep{chen2022unigeo}, and Geometry3K \citep{lu2021Geometry3K}, which offer a very narrow coverage of visual reasoning.

Later developments attempted to broaden the scope. MMMU \citep{yue2024mmmu} provides 505 college-level visual questions, but its complexity is limited by the use of multiple-choice format. MathVista \citep{lu2023mathvista} combines 28 existing and 3 new datasets, totaling 5k samples (1k test), although \citet{qiao2024we} noted issues with data quality.

The latest benchmarks face similar limitations. We-Math \citep{qiao2024we} includes 1.7k visual samples but again only uses the multiple-choice format. MathVerse \citep{zhang2024mathverse} and MATH-V \citep{wang2024measuring} both incorporate over 1.5k free-form solutions, but lack topic coverage due to their focus on simpler problems or high-school competition challenges.

\noindent
Our \umathvisual subset embraces the \textbf{free-form response format for visual problems} while adhering to the topics of \textbf{university coursework}.




\paragraph{Mathematical solution verification.} The open-ended nature of answers and ambiguity in mathematical expressions make evaluating math solutions particularly challenging. As a result, many benchmarks use multiple-choice questions for ease of grading, though this can simplify the tasks and offer hints that models can exploit \citep{li2024multiplechoicequestionsreallyuseful,pezeshkpour2023largelanguagemodelssensitivity}.

Free-form evaluation by LLM judges, while widespread \citep{zheng2023judging}, is prone to errors that are often overlooked and unaccounted for, compromising reliability \citep{zheng2023judging}. Therefore, tools allowing for assessment of automatic evaluators --- meta-evaluations --- are crucial. Recent studies also indicate that evaluating math solutions is challenging for LLMs \citep{mr-gsm8k, mr-math} and that judgment performance correlates with problem-solving performance without fully aligning with it \citep{stephan2024calculationadjudicationexaminingllm}, further reinforcing the relevance of meta-evaluations.

There are existing datasets suited for mathematical meta-evaluations: PRM800K \citep{lightman2023let} contains 800K annotated steps from 75K solutions to 12K MATH dataset problems, FELM \citep{zhao2024felm} provides GPT-3.5 annotations for solutions to 208 GSM8K and 194 MATH problems, MR-GSM8K \citep{mr-gsm8k} and MR-MATH \citep{mr-math} introduce meta-evaluation tasks based on the problems from GSM8K and MATH. These are all essentially based on GSM8K and MATH datasets, neglecting meta-evaluation for more advanced mathematical areas.

\noindent
Our \mumath benchmark is based on \umath problems, enabling \textbf{university-level meta-evaluations}.

\section{\umath} \label{sec:methodology}

We present \textbf{\umath} --- a benchmark of 1,100 problems designed to evaluate LLMs' proficiency in university-level mathematics. Following prior work~\citep{hendrycks2020measuring,hendrycks2021measuring,cobbe2021training,fang2024mathodyssey,yue2024mmmu}, we use \textbf{Accuracy} as our main performance metric, employing an LLM judge \citep{zheng2023judging} to test evaluated responses against the golden labels. A problem is only considered solved if each of the questions included with the problem statement is answered correctly and fully (e.g. if one of the questions asks to find the saddle points of a function, all of them have to be found).

\subsection{Dataset Curation} \label{sec:data}

We collaborate with \anonymized{Gradarius}, a platform providing math-specialized learning content and software for top US universities, sourcing tens of thousands of problems from ongoing courses across various institutions. Both problems and solutions are crafted by subject matter experts, representing real-world academic standards, and have not been externally published prior to our work. To build our benchmark, we select the most challenging problems available. In particular, we seek to filter out any calculation-intensive problems and focus on evaluating reasoning rather than arithmetical aptitude, as LLMs are not designed to perform arithmetic and are inherently prone to errors \citep{hendrycks2021measuring,lewkowycz2022solving}.

First, we filter out problems with short solutions ($<100$ characters), problems in multiple-choice format, and problems marked as implying calculator use. Additionally, for visual problems, we choose to keep only those containing a single image, for evaluation simplicity.

Next, we employ several small language models --- Llama-3.1 8B \citep{dubey2024llama}, Qwen2 7B \citep{yang2024qwen2technicalreport}, Mistral 7B \citep{jiang2023mistral7b}, Mathstral 7B, NuminaMath 7B \citep{numina_math_7b} --- to solve the problems and select 150 most challenging ones per subject, based on the average solution rate. By using a diverse set of model families, we avoid allowing any individual one to be overly influential in problem selection.

Lastly, we manually curate the selected problems using our in-house mathematical experts and the \anonymized{Gradarius} content team to ensure the absence of erroneous problem statements or golden labels.

Following the data curation, we enlist a team of academic experts from the \anonymized{Stevens Institute of Technology}, who actively teach various Calculus courses. These experts thoroughly review the problems to verify whether they are suitable for assessing the subject knowledge expected of university students. Overall, only 4.3\% of the problems are categorized as high-school rather than university-level.

\subsection{Dataset Statistics}

The \umath benchmark comprises \textbf{1,100} mathematical problems spanning \textbf{6 subjects}, with about \textbf{20\%} of the problems including visual elements (graphs, tables, geometric figures). Table~\ref{table:subjects} summarizes the problems' distribution across the subjects, together with the average number of questions posed and answers expected per problem (e.g. the task could be to find the local minima, maxima, and saddle points of a function, while the correct answer might contain no extrema and two saddle points).

\begin{table}[!ht]
    \centering
    \begin{adjustbox}{width=1.0\linewidth}
    \begin{tabular}{lcc|cccc}
    \toprule
    \textbf{Math Subject} & \textbf{\#Textual} & \textbf{\#Visual} & \textbf{Avg. Questions} & \textbf{Avg. Answers} \\
    \midrule
    Algebra                    & 150 & 30 & 1.93 & 1.28 \\
    Differential Calculus      & 150 & 70 & 2.37 & 1.15 \\
    Integral Calculus          & 150 & 58 & 1.09 & 1.01 \\
    Multivariable Calculus     & 150 & 28 & 1.74 & 1.09 \\
    Precalculus                & 150 & 10 & 1.51 & 1.23 \\
    Sequences and Series       & 150 & 4  & 1.36 & 1.00 \\
    \midrule
    All                        & 900 & 200 & 1.66 & 1.12 \\
    \bottomrule
    \end{tabular}
    \end{adjustbox}
    \caption{Statistics across \umath subjects: counts of text-only and visual problems, average questions per problem, and average answers per question}
    \label{table:subjects}
\end{table}

\subsection{Meta-Evaluation Framework (\mumath)} \label{subsec:meta-eval}

Evaluating mathematical problems is not straightforward, with even simple expressions such as $x \cdot 0.5$ having alternative valid forms such as $\frac{x}{2}$, $x \div 2$, $x / 2$, or unsimplified variants like $9x/18$. In practice, evaluating free-form solutions requires testing expression equivalence in much less trivial cases, especially with more advanced problems (see Appendix~\ref{appendix:meta-eval-tasks} for an example). To systematically study the ability of LLMs to evaluate free-form mathematical solutions on advanced university-level problems, we introduce the $\pmb{\mu}$\textbf{-MATH} benchmark. It consists of a curated subset of \umath samples, supplied with LLM-generated solutions, both correct and not. Four solutions are generated for each of the problems --- using Qwen2.5 72B, Llama-3.1 8B, GPT-4o and Gemini 1.5 Pro models. We focus on text-only problems due to the limited size of the \umathvisual subset.

Solution correctness is determined using a combination of manual labeling and automatic verification via \anonymized{Gradarius}-API, which allows to test formal equivalence of mathematical expressions. Whenever the API classifies an LLM-produced answer as coinciding with the golden label, we can be confident in that answer's correctness. However, a negative API response does not imply incorrectness, since extraction of the answer from the full solution and its subsequent conversion into an API-compatible expression format are imperfect. Hence, solutions with negative API responses, which occur roughly 40\% of the time, are labeled by in-house math experts, same as described in Section~\ref{sec:data}.

Our internal experts also review all the problems, including the ones with all the solutions auto-labeled, to assess their evaluation difficulty. In the end, we select \textbf{271 \umath problems} (around \textbf{25\%}) based on these difficulty estimates, resulting in a total of \textbf{1,084 samples}. The final set does not aim to reflect the overall \umath distribution, but rather provide a robust and challenging test for LLM judges.

A tested model is provided with a problem statement, a reference answer, and a solution to evaluate and is expected to produce a correctness judgment to be compared against the golden verdict. We treat this as a binary classification task, using the \textbf{macro-averaged F1-score} as our primary metric. To offer a finer-grained evaluation, we also report Positive Predictive Value (PPV or Precision) and True Positive Rate (TPR or Recall) for the positive class, as well as Negative Predictive Value (NPV) and True Negative Rate (TNR) for the negative class. We report scores calculated both overall (all samples) and per originating model, separately for each of the four author models.

\section{Experiments and Results} \label{sec:experiments}

\subsection{Experimental Setup} \label{subsec:setup}

\begin{table}[!htbp]
    \centering
    \begin{adjustbox}{width=1.0\linewidth}
    \begin{tabular}{llcccc}
        \toprule
        \textbf{Model} & \textbf{Source} & \textbf{Size(s)} & \textbf{Visual} & \textbf{Open-weights} & \textbf{Reasoner} \\
        \midrule
        Ministral 2410 & \citet{ministral_website_2024} & 8B & \xmark & \cmark & \xmark \\
        Mistral Small 2501 & \citet{mistralsmall3_website_2024} & 24B & \xmark & \cmark & \xmark \\
        Mistral Large 2411 & \citet{mistrallarge2_website_2024} & 123B & \xmark & \cmark & \xmark \\
        DeepSeek-V3 & \citet{DeepSeek-V3} & MoE 37/685B & \xmark & \cmark & \xmark \\
        Qwen2.5-Math & \citet{yang2024qwen25math} & 7B, 72B & \xmark & \cmark & \xmark \\
        Qwen2.5 & \citet{qwen2.5} & 7B, 32B, 72B & \xmark & \cmark & \xmark \\
        Athene-V2 Chat & \citet{athene_v2_website} & 72B & \xmark & \cmark & \xmark \\
        Llama-3.1 & \citet{dubey2024llama} & 8B, 70B & \xmark & \cmark & \xmark \\
        Llama-3.1 Nemotron & \citet{wang2024helpsteer2} & 70B & \xmark & \cmark & \xmark \\
        Llama-3.3 & \citet{wang2024helpsteer2} & 70B & \xmark & \cmark & \xmark \\
        
        Pixtral 12B 2409 & \citet{pixtral12b_website_2024} & 12B & \cmark & \cmark & \xmark \\
        Pixtral Large 2411 & \citet{pixtral12b_website_2024} & 124B & \cmark & \cmark & \xmark \\
        Qwen2-VL & \citet{yang2024qwen2technicalreport} & 7B, 72B & \cmark & \cmark & \xmark \\
        Llama-3.2 & \citet{llama_32_website} & 11B, 90B & \cmark & \cmark & \xmark \\[0.5em]

        Claude 3.5 Sonnet (new) & \citet{claude_35_sonnet_website_2024} & \small{unknown} & \cmark & \xmark & \xmark \\
        GPT-4o-mini-2024-07-18 & \citet{gpt4o_website_2024} & \small{unknown} & \cmark & \xmark & \xmark \\
        GPT-4o-2024-08-06 & \citet{gpt4o_website_2024} & \small{unknown} & \cmark & \xmark & \xmark \\
        Gemini 1.5 Flash 002 & \citet{geminiteam2024gemini15unlockingmultimodal} & \small{unknown} & \cmark & \xmark & \xmark \\
        Gemini 1.5 Pro 002 & \citet{geminiteam2024gemini15unlockingmultimodal} & \small{unknown} & \cmark & \xmark & \xmark \\[0.5em]

        DeepSeek-R1 & \citet{DeepSeek-R1} & MoE 37/685B & \xmark & \cmark & \cmark \\
        QwQ-Preview & \citet{qwq_website_2024} & 32B & \xmark & \cmark & \cmark \\
        QVQ-Preview & \citet{qvq_website_2024} & 72B & \cmark & \cmark & \cmark \\[0.5em]

        o1-mini-2024-09-12 & \citet{o1-mini_website_2024} & \small{unknown} & \xmark & \xmark & \cmark \\
        o3-mini-2025-01-31 & \citet{o3-mini_website_2024} & \small{unknown} & \xmark & \xmark & \cmark \\
        o1-2024-12-17 & \citet{o1_website_2024} & \small{unknown} & \cmark & \xmark & \cmark \\
        Gemini 2.0 Flash Thinking (exp-01-21) & \citet{gemini2-flash-thinking_website_2024} & \small{unknown} & \cmark & \xmark & \cmark \\
        \bottomrule
    \end{tabular}
    \end{adjustbox}
    \caption{The LLMs used in our work.}
    \label{table:llm_visual_open_source}
\end{table}

We select some of the recent top-performing LLMs to evaluate (Table~\ref{table:llm_visual_open_source}). All the non-reasoning models are restricted to a single generation of 4,096 tokens with temperature set to 0.

For reasoners, the token limit is 32,768. Note that o-series models do not allow for inference temperature control, always having a default nonzero temperature. Our internal tests on a subset of the models, including DeepSeek-R1 and QwQ-32B-Preview for the reasoner subset, show negligible differences in accuracy under greedy decoding and four-rollout Pass@1 with a temperature of 0.6 (average accuracy over four independent launches), nor do we observe any significant variation across the rollouts. We thus adhere to a single-generation scheme for reasoners as well, employing greedy decoding for all the models except the o-series.

We use chain-of-thought prompting \citep{wei2022chain} with the prompt provided in Appendix~\ref{appendix:prediction} and o3-mini as a judge, due to the model being simultaneously one of the most performant and balanced judges according to our meta-evaluations (see Section~\ref{subsec:meta-eval-exps}), as well as cost-effective and widely available, allowing for easier reproduction.

\subsection{\umath Results}

\begin{table*}[!htb]
    \centering
    \begin{adjustbox}{width=\linewidth}
    \begin{tabular}{lccc||cc|cc|cc|cc|cc|cc}
        \toprule
        \multirow{3}{*}{\textbf{Model}} & \multirow{3}{*}{\textbf{\umath}} &\multicolumn{2}{c}{\textbf{\umath}} & \multicolumn{2}{c}{\textbf{Algebra}} & \multicolumn{2}{c}{\textbf{Diff. C.}} & \multicolumn{2}{c}{\textbf{Integral C.}} & \multicolumn{2}{c}{\textbf{Multivar C.}} & \multicolumn{2}{c}{\textbf{Precalculus}} & \multicolumn{2}{c}{\textbf{Seq.\& Series}} \\
        \textbf{} & \textbf{} & \textbf{T} & \textbf{V} & \textbf{T} & \textbf{V} & \textbf{T} & \textbf{V} & \textbf{T} & \textbf{V} & \textbf{T} & \textbf{V*} & \textbf{T} & \textbf{V*} & \textbf{T} & \textbf{V*} \\[-4pt]
        \tiny  & \tiny  & \tiny 900 & \tiny 200 & \tiny 150 & \tiny 30 & \tiny 150 & \tiny 70 & \tiny 150 & \tiny 58 & \tiny 150 & \tiny 28 & \tiny 150 & \tiny 10 & \tiny 150 & \tiny 4 \\[-2pt]
        \midrule
        \multicolumn{16}{c}{\textbf{Text-only models}} \\
        \midrule   
        Ministral 8B & 23.1 & 26.9 & 6.0 & 60.0 & 6.7 & 13.3 & 8.6 & 10.0 & 5.2 & 12.7 & 3.6 & 47.3 & 0.0 & 18.0 & 0.0  \\
        Llama-3.1 8B & 29.5 & 33.7 & 11.0 & 60.0 & 3.3 & 17.3 & 10.0 & 22.7 & 19.0 & 23.3 & 3.6 & 50.7 & 20.0 & 28.0 & 0.0  \\
        Qwen2.5 7B & 43.3 & 50.4 & 11.0 & 86.0 & 20.0 & 30.7 & 4.3 & 32.0 & 19.0 & 36.7 & 3.6 & 78.7 & 10.0 & 38.7 & 0.0  \\
        Qwen2.5-Math 7B & 45.5 & 53.0 & 11.5 & 84.7 & 6.7 & 32.0 & 8.6 & 24.0 & 17.2 & 44.0 & 10.7 & 81.3 & 0.0 & 52.0 & 50.0  \\[0.5em]

        Mistral Small (24B) & 34.8 & 39.9 & 12.0 & 80.7 & 13.3 & 13.3 & 10.0 & 13.3 & 15.5 & 25.3 & 14.3 & 70.7 & 0.0 & 36.0 & 0.0  \\
        Qwen2.5 32B & 52.4 & 60.4 & 16.0 & 92.0 & 13.3 & 42.7 & 11.4 & 34.7 & 25.9 & 50.0 & 17.9 & 85.3 & 0.0 & 58.0 & 0.0  \\[0.5em]

        Llama-3.1 70B & 35.2 & 40.4 & 11.5 & 79.3 & 3.3 & 17.3 & 17.1 & 16.0 & 10.3 & 26.7 & 7.1 & 68.0 & 0.0 & 35.3 & 50.0  \\
        Llama-3.1 Nemotron 70B & 42.5 & 47.7 & 19.5 & 84.0 & \textbf{23.3} & 29.3 & 21.4 & 21.3 & 19.0 & 40.7 & 14.3 & 67.3 & 20.0 & 43.3 & 0.0  \\
        Llama-3.3 70B & 44.7 & 51.7 & 13.5 & 83.3 & 6.7 & 35.3 & 11.4 & 27.3 & 20.7 & 48.7 & 10.7 & 68.7 & 10.0 & 46.7 & 25.0  \\
        Qwen2.5 72B & 51.2 & 58.9 & 16.5 & 90.7 & 16.7 & 36.7 & 15.7 & 35.3 & 17.2 & 52.0 & 14.3 & 84.0 & 10.0 & 54.7 & 50.0  \\
        Athene-V2 Chat (72B) & 54.9 & 62.9 & 19.0 & 87.3 & 10.0 & 43.3 & 22.9 & 36.7 & 17.2 & 62.0 & 21.4 & \textbf{90.7} & 0.0 & 57.3 & \textbf{75.0}  \\
        Qwen2.5-Math 72B & 59.5 & 68.7 & 18.0 & 94.7 & 6.7 & 46.0 & 12.9 & \textbf{44.0} & 25.9 & \textbf{69.3} & 21.4 & 89.3 & 10.0 & 68.7 & \textbf{75.0}  \\[0.5em]

        Mistral Large (123B) & 47.6 & 55.6 & 12.0 & 85.3 & 13.3 & 32.0 & 8.6 & 36.7 & 15.5 & 45.3 & 14.3 & 78.0 & 0.0 & 56.0 & 25.0  \\
        DeepSeek-V3 (MoE 37/685B) & \textbf{62.6} & \textbf{69.3} & \textbf{32.5} & \textbf{96.0} & 10.0 & \textbf{49.3} & \textbf{30.0} & 38.7 & \textbf{39.7} & \textbf{69.3} & \textbf{42.9} & 90.0 & \textbf{40.0} & \textbf{72.7} & 50.0  \\

        \midrule
        \multicolumn{16}{c}{\textbf{Multimodal models}} \\
        \midrule   
        
        Pixtral 12B & 17.5 & 17.9 & 16.0 & 40.0 & 23.3 & 10.7 & 30.0 & 4.7 & 3.4 & 6.7 & 7.1 & 32.0 & 0.0 & 13.3 & 0.0  \\
        Llama-3.2 11B & 20.4 & 22.9 & 9.0 & 52.0 & 3.3 & 7.3 & 20.0 & 1.3 & 3.4 & 13.3 & 0.0 & 44.0 & 10.0 & 19.3 & 0.0  \\
        Qwen2-VL 7B & 26.3 & 27.1 & 22.5 & 58.7 & 10.0 & 18.7 & 37.1 & 11.3 & 17.2 & 14.0 & 17.9 & 42.7 & 10.0 & 17.3 & 0.0  \\[0.5em]
        
        Llama-3.2 90B & 37.2 & 41.8 & 16.5 & 82.0 & 23.3 & 21.3 & 27.1 & 11.3 & 5.2 & 30.0 & 10.7 & 70.0 & 0.0 & 36.0 & 25.0  \\
        Qwen2-VL 72B & 41.8 & 43.9 & 32.5 & 80.0 & 26.7 & 29.3 & 44.3 & 22.0 & 27.6 & 32.0 & 28.6 & 66.0 & 10.0 & 34.0 & 25.0  \\
        Pixtral Large (124B) & 47.8 & 51.4 & 31.5 & 82.7 & 33.3 & 30.0 & 32.9 & 24.7 & \textbf{32.8} & 46.7 & 28.6 & 73.3 & 30.0 & 51.3 & 0.0  \\[0.5em]
        
        Claude Sonnet 3.5 & 38.7 & 40.7 & 30.0 & 75.3 & 30.0 & 20.7 & 41.4 & 12.0 & 15.5 & 33.3 & 39.3 & 64.0 & 20.0 & 38.7 & 0.0  \\
        GPT-4o-mini & 43.4 & 47.2 & 26.0 & 87.3 & 13.3 & 26.0 & 32.9 & 16.7 & 17.2 & 37.3 & 39.3 & 76.0 & 20.0 & 40.0 & 50.0  \\
        GPT-4o & 50.2 & 53.9 & 33.5 & 90.0 & 33.3 & 30.0 & 37.1 & 27.3 & 27.6 & 49.3 & 42.9 & 80.0 & 30.0 & 46.7 & 0.0  \\
        Gemini 1.5 Flash & 57.8 & 61.2 & 42.5 & 90.7 & 46.7 & 47.3 & 47.1 & 30.7 & 31.0 & 55.3 & 53.6 & 82.7 & 30.0 & 60.7 & 50.0  \\
        Gemini 1.5 Pro & \textbf{67.2} & \textbf{71.7} & \textbf{47.0} & \textbf{92.0} & \textbf{60.0} & \textbf{62.0} & \textbf{50.0} & \textbf{47.3} & 27.6 & \textbf{65.3} & \textbf{60.7} & \textbf{90.0} & \textbf{50.0} & \textbf{73.3} & \textbf{75.0}  \\
    
        \midrule
        \multicolumn{16}{c}{\textbf{Reasoning models}} \\
        \midrule   

        QVQ-72B-Preview & 65.0 & 69.7 & 44.0 & 94.0 & 33.3 & 54.0 & 41.4 & 41.3 & 55.2 & 65.3 & 50.0 & 95.3 & 30.0 & 68.0 & 0.0  \\
        QwQ-32B-Preview & 73.1 & 82.7 & 30.0 & 95.3 & 3.3 & 70.0 & 24.3 & 67.3 & 50.0 & 80.7 & 32.1 & 97.3 & 20.0 & 85.3 & 50.0  \\
        DeepSeek-R1 (MoE 37/685B) & 80.7 & 91.3 & 33.0 & 96.7 & 16.7 & 85.3 & 22.9 & 87.3 & 50.0 & 86.7 & 42.9 & 98.7 & 10.0 & 93.3 & \textbf{75.0}  \\[0.5em]
        
        o1-mini & 76.3 & 82.9 & 46.5 & 97.3 & 40.0 & 75.3 & 52.9 & 72.0 & 46.6 & 78.7 & 42.9 & 96.7 & 30.0 & 77.3 & 50.0  \\
    
        Gemini 2.0 Flash Thinking & 83.6 & 89.2 & \textbf{58.5} & 95.3 & \textbf{60.0} & 80.7 & 48.6 & 88.7 & \textbf{65.5} & 85.3 & \textbf{75.0} & 95.3 & \textbf{50.0} & 90.0 & 25.0  \\
        o3-mini & 82.2 & 92.8 & 34.5 & \textbf{99.3} & 10.0 & \textbf{88.0} & 17.1 & \textbf{90.7} & 60.3 & 85.3 & 50.0 & \textbf{99.3} & 20.0 & \textbf{94.0} & \textbf{75.0}  \\
        o1 & \textbf{86.8} & \textbf{93.1} & \textbf{58.5} & 97.3 & 50.0 & 86.0 & \textbf{57.1} & \textbf{90.7} & 63.8 & \textbf{92.0} & 60.7 & \textbf{99.3} & \textbf{50.0} & 93.3 & \textbf{75.0}  \\
        \bottomrule
    \end{tabular}
    \end{adjustbox}
    \caption{Comparison of models' results on \umath. Scores for various subjects are displayed along with the integral scores. T denotes accuracy over text-only tasks, V denotes accuracy over visual tasks. Asterisk denotes a small number of samples ($<30$). Images are not included in the prompt for text-only models, only the problem statements. Note that text-only models can solve a percentage of visual problems, due to either guessing, some of the problems being solvable without the accompanying images, or judgment errors discussed in Section \ref{subsec:meta-eval-exps}. \textbf{Bold} indicates the best result in each group.}
    \label{table:models_our_benchmark}
\end{table*}

Table~\ref{table:models_our_benchmark} summarizes the results of our experiments. We observe several key trends. 

\vspace{0.3em}
\noindent
\textbf{Reasoners offer breakthrough performance:} Reasoning models attain the top \umath, \umatht and \umathv scores of 86.8\%, 93.1\% and 58.5\% respectively, compared to 67.2\%, 71.7\% and 47.0\% for the standard-inference models.

\vspace{0.3em}
\noindent
\textbf{Open models are catching up on text-only problems, with DeepSeek in the lead:} DeepSeek-V3 achieves a \umatht score of 69.3\%, closely trailing the leading Gemini 1.5 Pro model with 71.7\%. DeepSeek-R1 (91.3\%) is only marginally behind o1, the best-performing reasoner (93.1\%).

\vspace{0.3em}
\noindent
\textbf{Open models lag behind in visual problems, where Gemini dominates:} The open-proprietary gap becomes much more pronounced when considering \umathv. In each `capability group' (smaller, larger, and reasoning models) the best open-weight result comes from the Qwen family (Qwen2-VL 7B: 27.1\%, Qwen2-VL 72B: 43.9\%, QVQ-72B-Preview: 44.0\%), trailing far behind Gemini models. Gemini leads the proprietary category across all scales with considerable margins (Gemini 1.5 Flash: 42.5\%, Gemini 1.5 Pro: 47.0\%, Gemini 2.0 Flash Thinking: 58.5\%).

\vspace{0.3em}
\noindent
\textbf{Visual comprehension is challenging:} \umathv scores are consistently much lower compared to \umatht, although manual examinations do not suggest the underlying problems to be any harder. Besides, transitioning from text-only to visual often causes degradation in models' textual performance: 48.1\% $\Rightarrow$ 42.9\% with Mistral and Pixtral Large, 26.1\% $\Rightarrow$ 18.6\% with smaller Llama-3.1 and Llama-3.2, 71.8\% $\Rightarrow$ 59.3\% with QwQ and QVQ Preview.

\vspace{0.8em}
\noindent
\textbf{Specialization trumps Size:} Larger models expectedly outperform smaller ones, but small-scale specialists like Qwen2.5-Math 7B can surpass models 10 times their size, such as Llama-3.1 70B. Similarly, Qwen2.5-Math 72B performs on par with a 685B mixture-of-experts DeepSeek-V3.

\vspace{0.3em}
\noindent
\textbf{Continuous Finetuning enhances performance:} Llama-3.1 70B $\Rightarrow$ Llama-3.1 Nemotron 70B and Qwen2.5-72B $\Rightarrow$ Athene-V2 72B yield 2.9\% and 5.2\% higher \umath accuracy respectively, suggesting that standard-inference models may not be fully optimized for their size and could use high-quality post-training data to improve further.
\vspace{3.745em}




\subsection{Meta-Evaluation (\mumath) Results} \label{subsec:meta-eval-exps}
Meta-evaluations follow the setup in Section~\ref{subsec:setup}.  Additionally, we experiment with two distinct prompting schemes --- a standard Automatic Chain-of-Thought (AutoCoT) prompt involving a simple task description followed by an instruction to think step-by-step, and a manual Chain-of-Thought prompt (which we refer to as simply CoT) with explicit instructions on which steps to follow --- finding the latter performs best and using it as our default. The judge's output is further processed by an extractor model (Qwen2.5 72B is fixed for consistency), prompted to produce a single label --- `Yes', `No' or `Inconclusive' --- with `Inconclusive' reserved for refusals or generation failures and treated as incorrect. Reference Appendix~\ref{appendix:judgment} for the full prompt contents. The main results are presented in Table~\ref{table:mu-math_results}. We summarize our conclusions in the following.

\begin{table*}[!htbp]
    \centering
    \begin{adjustbox}{width=1.0\linewidth}
    \begin{tabular}{lc||c|cccc|cccc}
    \toprule
        \multirow{2}{*}{\textbf{Model}} & 
        \multirow{2}{*}{\textbf{\umathtext}} &
        \multicolumn{5}{c}{\textbf{\mumath}} & 
        \multicolumn{1}{c}{\textbf{\mumath}$_\text{Qwen}$} & 
        \multicolumn{1}{c}{\textbf{\mumath}$_\text{Llama}$} &
        \multicolumn{1}{c}{\textbf{\mumath}$_\text{GPT}$} &
        \multicolumn{1}{c}{\textbf{\mumath}$_\text{Gemini}$} \\
    \cmidrule{3-11}
    &
    &
    \textbf{F1$_\text{CoT}$ \gray{/ F1$_\text{AutoCoT}$}} & \textbf{TPR} & \textbf{TNR} & \textbf{PPV} & \textbf{NPV} &
    \textbf{F1$_\text{CoT}$ \gray{/ F1$_\text{AutoCoT}$}} & 
    \textbf{F1$_\text{CoT}$ \gray{/ F1$_\text{AutoCoT}$}} & 
    \textbf{F1$_\text{CoT}$ \gray{/ F1$_\text{AutoCoT}$}} & 
    \textbf{F1$_\text{CoT}$ \gray{/ F1$_\text{AutoCoT}$}} \\
    \midrule

    Llama-3.1 8B & 33.7 & 52.0 \gray{/ 53.1} & 48.7 & 55.9 & 56.0 & 48.5 & 48.7 \gray{/ 49.6} & 49.2 \gray{/ 51.2} & 51.2 \gray{/ 57.6} & 55.5 \gray{/ 50.5} \\
    Ministral 8B & 26.9 & 60.5 \gray{/ 58.9} & 55.9 & 65.8 & 65.4 & 56.4 & 52.8 \gray{/ 55.7} & 63.1 \gray{/ 58.2} & 62.9 \gray{/ 60.9} & 58.3 \gray{/ 54.1} \\
    Qwen2.5-Math 7B & 53.0 & 61.9 \gray{/ 61.2} & 76.6 & 47.9 & 62.9 & 63.9 & 59.7 \gray{/ 56.7} & 63.8 \gray{/ 64.0} & 57.2 \gray{/ 58.5} & 63.8 \gray{/ 61.2} \\
    Qwen2.5 7B & 50.4 & 69.3 \gray{/ 67.0} & \textbf{78.7} & 59.8 & 69.3 & 70.8 & 62.4 \gray{/ 60.5} & 72.3 \gray{/ 72.4} & 68.3 \gray{/ 66.4} & 69.1 \gray{/ \textbf{65.0}} \\[0.5em]
    
    GPT-4o-mini & 47.2 & 72.3 \gray{/ \textbf{69.2}} & 59.0 & 88.1 & 85.1 & 65.1 & 69.3 \gray{/ 61.7} & 76.2 \gray{/ \textbf{78.5}} & \textbf{70.4} \gray{/ \textbf{69.8}} & 69.6 \gray{/ 64.3} \\
    Gemini 1.5 Flash & \textbf{61.2} & \textbf{74.8} \gray{/ 65.3} & 63.3 & \textbf{88.3} & \textbf{86.2} & \textbf{67.6} & \textbf{71.2} \gray{/ \textbf{61.9}} & \textbf{80.6} \gray{/ 70.8} & 70.1 \gray{/ 65.3} & \textbf{73.9} \gray{/ 59.7} \\
    
    \midrule

    Llama-3.1-70B & 40.4 & 61.0 \gray{/ 68.2} & 62.5 & 59.6 & 64.1 & 57.9 & 56.0 \gray{/ 63.8} & 57.0 \gray{/ 70.2} & 69.4 \gray{/ 69.8} & 58.8 \gray{/ 64.4} \\
    Qwen2.5-Math 72B & 68.7 & 74.0 \gray{/ 75.5} & \textbf{80.9} & 66.8 & 73.8 & 75.2 & 69.3 \gray{/ 68.8} & 77.3 \gray{/ 79.8} & 68.2 \gray{/ 69.2} & 76.8 \gray{/ 80.4} \\
    Qwen2.5 72B & 58.9 & 75.6 \gray{/ 75.1} & 77.1 & 74.2 & 77.5 & 73.7 & 70.5 \gray{/ 68.9} & 79.3 \gray{/ 80.1} & 73.7 \gray{/ 73.4} & 74.2 \gray{/ 73.8} \\
    Mistral Large & 55.6 & 76.6 \gray{/ 74.5} & 75.7 & 77.7 & 79.7 & 73.5 & 72.5 \gray{/ 70.8} & 78.6 \gray{/ 77.7} & 76.0 \gray{/ 74.4} & 75.0 \gray{/ 71.0} \\
    DeepSeek-V3 & 69.3 & 80.6 \gray{/ \textbf{81.5}} & 77.0 & 84.7 & 85.0 & 76.6 & \textbf{81.8} \gray{/ \textbf{76.0}} & 81.2 \gray{/ \textbf{86.2}} & 74.9 \gray{/ \textbf{80.1}} & 80.4 \gray{/ \textbf{82.7}} \\[0.5em]

    Claude 3.5 Sonnet & 40.7 & 74.8 \gray{/ 68.1} & 62.5 & \textbf{89.5} & \textbf{87.3} & 67.4 & 70.8 \gray{/ 64.1} & 77.9 \gray{/ 71.8} & 72.2 \gray{/ 68.1} & 73.8 \gray{/ 63.4} \\
    GPT-4o & 53.9 & 77.4 \gray{/ 74.2} & 70.1 & 85.9 & 85.1 & 71.3 & 74.2 \gray{/ 68.2} & 81.8 \gray{/ 78.9} & 77.5 \gray{/ 75.8} & 72.6 \gray{/ 70.5} \\
    Gemini 1.5 Pro & \textbf{71.7} & \textbf{81.5} \gray{/ 69.8} & 78.5 & 84.7 & 85.2 & \textbf{78.2} & 78.9 \gray{/ 65.4} & \textbf{83.6} \gray{/ 74.8} & \textbf{79.3} \gray{/ 69.1} & \textbf{80.5} \gray{/ 65.8} \\

    \midrule

    QwQ-32B-Preview & 82.7 & 81.0 \gray{/ 79.6} & 85.7 & 75.9 & 80.5 & 82.2 & 81.9 \gray{/ 77.8} & 81.3 \gray{/ 79.4} & 76.1 \gray{/ 76.8} & 80.8 \gray{/ 79.8} \\
    DeepSeek-R1 & 91.3 & 84.3 \gray{/ 83.8} & 77.3 & \textbf{92.2} & \textbf{91.7} & 78.4 & 80.8 \gray{/ 81.1} & 87.1 \gray{/ 85.8} & 81.8 \gray{/ 81.5} & 84.7 \gray{/ 83.4} \\[0.5em]

    Gemini 2.0 Flash-Thinking & 89.2 & 80.2 \gray{/ 81.2} & 89.2 & 70.8 & 77.4 & 85.4 & 77.3 \gray{/ 78.0} & 81.1 \gray{/ 84.0} & 76.1 \gray{/ 78.9} & 82.6 \gray{/ 79.4} \\
    o1-mini & 82.9 & 83.4 \gray{/ 84.3} & 78.5 & 88.8 & 88.8 & 78.7 & 80.0 \gray{/ 83.8} & 88.0 \gray{/ 87.0} & 81.1 \gray{/ 82.2} & 81.3 \gray{/ 80.8} \\
    o3-mini & 92.8 & 89.6 \gray{/ 89.8} & 89.0 & 90.2 & 91.1 & 88.0 & 87.7 \gray{/ \textbf{88.4}} & 93.2 \gray{/ 93.6} & 88.2 \gray{/ 88.6} & 86.7 \gray{/ 85.7} \\
    o1 & \textbf{93.1} & \textbf{90.1} \gray{/ \textbf{90.2}} & \textbf{91.4} & 88.6 & 90.0 & \textbf{90.2} & \textbf{86.1} \gray{/ 85.7} & \textbf{94.4} \gray{/ \textbf{94.7}} & \textbf{88.9} \gray{/ \textbf{89.3}} & \textbf{88.7} \gray{/ \textbf{89.1}} \\

    \bottomrule
    \end{tabular}
    \end{adjustbox}
    \caption{Judgment performance on \mumath benchmark using CoT prompting; Macro F1-score (F1), True Positive Rate (TPR), True Negative Rate (TNR), Positive Predictive Value (PPV) and Negative Predictive Value (NPV) are presented, with F1 as the primary metric. The second number within each F1 column written in gray represents the score under AutoCoT prompting. \mumath columns display integral scores over the entire benchmark, while \mumath$_\text{<model>}$ columns denote subsets with solutions generated by specific author models. \umathtext accuracy is added for comparison of each model's performance as a problem-solver vs. as a judge. \textbf{Bold} indicates the best result in each column.} 
    \label{table:mu-math_results}
\end{table*}

\begin{figure*}[!ht]
    \centering
    \includegraphics[width=1.0\textwidth]{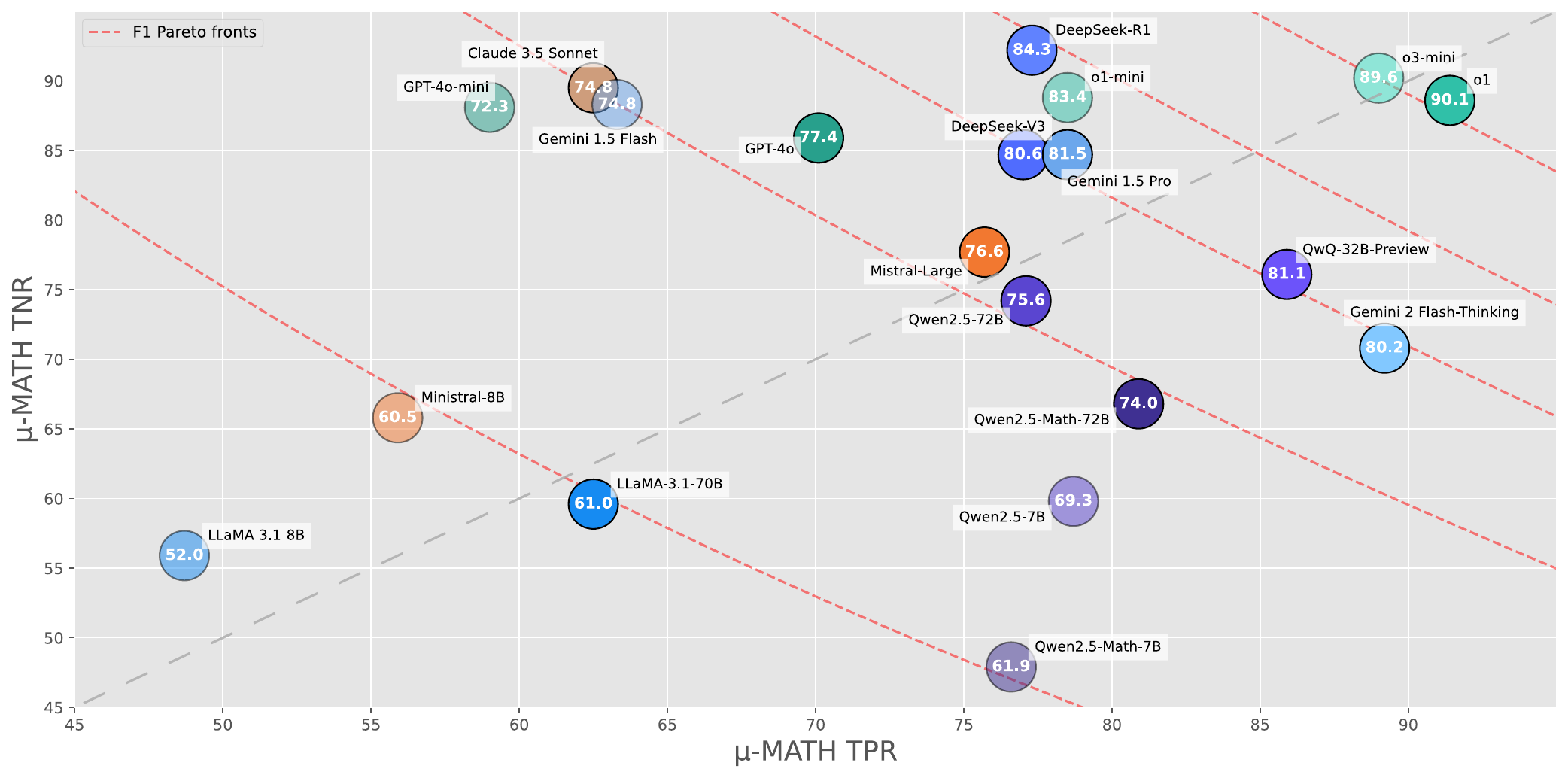}
    \caption{True Positive Rate vs True Negative Rate of judges on \mumath. The value inside of the marker denotes the F1-score.}
    \label{fig:mumath_tpr-tnr}
\end{figure*}

\begin{figure*}[!ht]
    \centering
    \includegraphics[width=1.\textwidth]{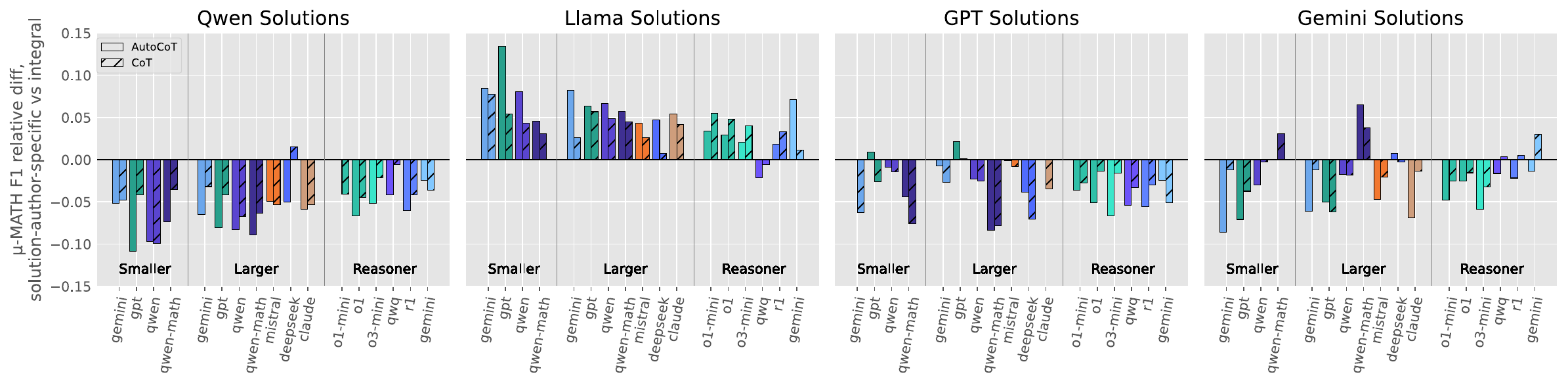}
    \caption{Relative difference in judge \mumath F1 scores: performance on a specific author's solutions vs. overall performance. Each pane corresponds to one of the author models. X-axis specifies the judge model (in three groups: small, large, reasoner). Bar pairs compare the difference for AutoCoT vs. manual CoT prompting. The three least performant models (Ministral 8B, Llama-3.1-8B and -70B) are excluded due to outlier behavior (e.g. Appendix~\ref{appendix:mumath_inconclusive}).}
    \label{fig:mumath_prompts_bias}
\end{figure*}

\vspace{0.3em}
\noindent
\textbf{Judgment is non-trivial:} In non-reasoners, the maximum attainable F1 score is only 81.5\%, and while reasoning models offer significant improvements, reaching a high F1 mark of 90.1\%, our results underscore that LLM judges remain fallible --- even when applied in an objective domain with access to ground truth labels and using the best current systems. This observation is important because judges' error rates directly limit evaluation precision. Moreover, in cases where judgment errors are systematic in nature as opposed to pure noise --- an issue we explore later with an example --- this cannot be overcome with sheer data volume.

\vspace{0.3em}
\noindent
\textbf{Judgment is distinct from problem-solving:} Superior problem-solving does not necessarily translate to better judgment, as illustrated, for instance, with Qwen2.5 vs. Qwen2.5-Math scores. In fact, our results suggest a trade-off between these skills, tracing to reasoning-coherence tradeoff and manifesting in judges' behavioral differences. These are most apparent (Figure~\ref{fig:mumath_tpr-tnr}) in non-reasoners: proprietary models tend towards conservatism (relatively high TNR compared to TPR), whereas Qwen models, particularly math specialists, exhibit the opposite. See Appendix~\ref{appendix:mumath_vs_umath} for more detailed discussion.

\vspace{0.3em}
\noindent
\textbf{Reasoners exceed the Pareto frontier:} Reasoning models improve substantially in both problem-solving and judgment performance over the previous model generation. Notably, the two best performing systems, o1 and o3-mini, are also among the most balanced with respect to TPR-TNR parity.

\vspace{0.3em}
\noindent
\textbf{Prompting effects are substantial yet inhomogeneous across models:} In non-reasoners, switching from AutoCoT to CoT generally maintains or improves judgment performance and reduces author bias (see paragraph below), except for Llama models, which suffer an increase in inconclusive judgments (Appendix~\ref{appendix:mumath_inconclusive}, Table~\ref{table:mu-math_results}). Gemini 1.5 models benefit the most (>10\% F1 gain), becoming the top non-reasoners and surpassing the Qwen, DeepSeek, and GPT models that beat Gemini with AutoCoT. Reasoner systems, however, remain largely unaffected by the change in prompting. 

\vspace{0.3em}
\noindent
\textbf{Judges exhibit model-specific biases:} We observe a consistent trend toward better performance on Llama solutions and worse performance on Qwen solutions (see Figure~\ref{fig:mumath_prompts_bias}). The author bias is most pronounced with smaller judges under AutoCoT prompting and reduced when moving toward more capable models and switching to CoT in the case of non-reasoners. At the same time, no noticeable self-judgment effects are observed.

\section{Conclusion} \label{sec:conclusion}
We introduce \textbf{\umath}, a novel multi-modal benchmark for university-level mathematical reasoning, featuring 1,100 unpublished problems sourced from real teaching materials spanning six university subjects, with 20\% involving visual elements. In addition, we provide \textbf{\mumath}, a \umath-derived meta-evaluation dataset enabling rigorous assessment of LLM judges.

Our experiments reveal LLM weaknesses in advanced mathematical reasoning, particularly visual tasks (achieving 58.5\% accuracy vs. 93.1\% for text-only). Enabling visual reasoning is difficult, often degrading textual performance, and is underdeveloped --- especially in open-weight models, which lag significantly behind proprietary ones despite near parity in text-only problems. Nevertheless, continuous fine-tuning, reasoning-first training, and mathematical specialization boost performance, suggesting considerable growth potential.

Judgment proves both distinct from problem-solving and non-trivial for LLMs, with only the most capable models attaining meaningfully high performance while still peaking at an imperfect 90.1\% F1-score mark. Additionally, we discover pronounced biases and instabilities in judgment performance as well as distinctive behavioral patterns, underscoring the utility and necessity of meta-evaluations.


\newpage
\section*{Limitations} 
While \umath offers a diverse set of university curricula problems, it does not cover the full range of advanced  mathematical subjects. In addition, while the textual parts of our benchmarks demonstrate good model separability across the broad spectrum of recent models, these parts start to approach saturation with the reasoning systems, further necessitating expansion into more advanced topics such as, for example, complex analysis. Moreover, the 20\% fraction of visual problems, while reflective of real-world coursework, limits the scope of visual reasoning evaluations. Furthermore, visual problems are not covered by our meta-evaluations.

\noindent
Although accuracy is a standard metric of choice, it discards a lot of signal and does not allow for finer-grained analyses. Furthermore, reliance on LLM judges introduces errors and biases, and while we do quantify these to some extent, that is only a first step, and additional mitigation mechanisms would need to be put in place in order to account for the errors in a principled manner. 

\paragraph{Future Work.} Future research can focus on the design of assessment protocols that allow partial credit to enable finer-grained problem-solving evaluations. Another important direction is bridging the gap between quantifying the uncertainty and bias induced by auto-evaluations and controlling for them. Finally, a possible way of overcoming saturation, apart from going through a costly process of curating new data, is coming up with adversarial task creation or modification approaches, which we see as particularly relevant for meta-evaluations. By open-sourcing our data and evaluation code, we strive to facilitate further research and encourage development of models better equipped for complex, real-world mathematical problems.






\section*{Ethics Statement}

We collected all data in \umath and \mumath with appropriate permissions, ensuring no personal or proprietary information is included. The datasets consist solely of mathematical problems and solutions, without any sensitive content. We open-sourced the datasets and code under suitable licenses to support transparency and research advancement. There are no known conflicts of interest associated with this work.

\section*{Reproducibility Statement}

All datasets and evaluation code will be available on GitHub. Detailed descriptions of data collection and processing are presented in Section~\ref{sec:methodology}. The experimental setup, including model configurations and prompts, is described in Section~\ref{sec:experiments}, with the full prompts provided in Appendices~\ref{appendix:prediction} and \ref{appendix:judgment}.

\FloatBarrier
\bibliography{custom}

\FloatBarrier
\nolinenumbers
\onecolumn
\appendix

\FloatBarrier
\newpage
\section{Problem examples}
\subsection{\umath Sample Problems} \label{appendix:umath-problems}

\begin{figure*}[htbp!]
    \centering
    \autogrid{
        \examplebox{Example 1: Algebra}{
        Write a logarithmic equation corresponding to the graph, using log base $3$:\\\includegraphics[width=0.5\linewidth]{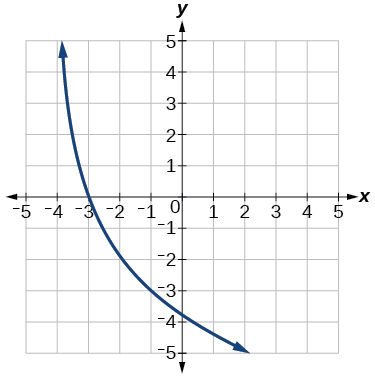}
        \examplesplit
        $$-3 \cdot \log_{3}(x+4)$$
        },
        \examplebox{Example 2: Integral Calculus}{
        Solve the integral:\\
        $$
        \int \frac{ -9 \cdot \sqrt[3]{x} }{ 9 \cdot \sqrt[3]{x^2} + 3 \cdot \sqrt{x} } \, dx
        $$
        \examplesplit
        \begin{flalign*}
            &-\frac{2}{27} \cdot \ln\left(\dfrac{\left|1 + 3 \cdot \sqrt[6]{x}\right|}{3} \right) - && \\
            &-\frac{1}{3} \sqrt[6]{x^2} - \frac{3}{2} \sqrt[6]{x^4} + \frac{2}{3} \sqrt[6]{x^3} + \frac{2}{9} \sqrt[6]{x} + C &&
        \end{flalign*}
        },
        \examplebox{Example 3: Precalculus Review}{
        Find a formula for the plotted sinusoidal function:\\\includegraphics[width=0.7\linewidth]{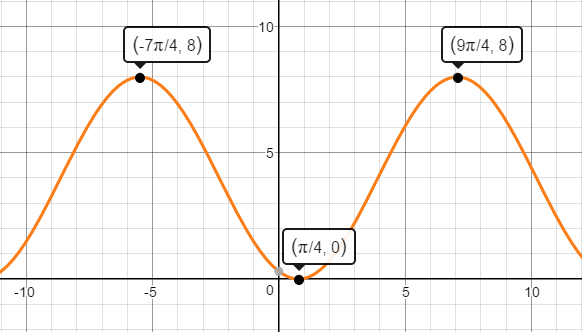}
        \examplesplit
        $$f(x) = -4\cdot\cos\left(\frac{1}{2}\cdot\left(x-\frac{\pi}{4}\right)\right)+4$$
        },
        \examplebox{Example 4: Multivariable Calculus}{
        $E$ is located inside the cylinder $x^2+y^2=1$ and between the circular paraboloids $z=1-x^2-y^2$ and $z=x^2+y^2$. Find the volume of $E$.
        \examplesplit
        $$\pi / 4$$
        },
        \examplebox{Example 5: Multivariable Calculus}{
        The graph of the polar rectangular region $D$ is given. Express the region $D$ in polar coordinates:\\\includegraphics[width=0.6\linewidth]{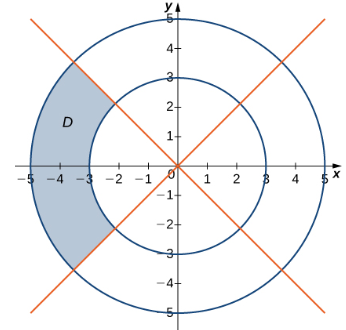}
        \examplesplit
        1. The interval of $r$ is $[3,5]$\\[1em]
        2. The interval of $\theta$ is $\left[\frac{3}{4}\cdot\pi,\frac{5}{4}\cdot\pi\right]$
        },
        \examplebox{Example 6: Differential Calculus}{
        Sketch the curve:\\
        $$
        y = \frac{x^3}{6 \cdot (x+3)^2}
        $$
        Provide the following:\\
        \scriptsize{
        1. The domain (in interval notation) \\ 
        2. Vertical asymptotes \\ 
        3. Horizontal asymptotes \\ 
        4. Slant asymptotes \\ 
        5. Intervals where the function is increasing \\ 
        6. Intervals where the function is decreasing \\ 
        7. Intervals where the function is concave up \\ 
        8. Intervals where the function is concave down \\ 
        9. Points of inflection 
        }
        \examplesplit
        1. The domain: $(-\inf,-3) \cup (-3,\inf)$\\[0.2em]
        ... 
        }
    }
    \caption{Examples of text-only and visual problems from the \umath benchmark along with their subjects and golden answers.}
\end{figure*}

\FloatBarrier\newpage
\subsection{\umath Sample Problem with Reference Solution} \label{appendix:umath-problems-solutions}

\begin{figure*}[!htb]
    \centering
        \examplebox{Example: Differential Calculus}{
        \textbf{\umath Problem:} \\
            The function $s(t) = 2 \cdot t^3 - 3 \cdot t^2 - 12 \cdot t + 8$ represents the position of a particle traveling along a horizontal line.

            1. Find the velocity and acceleration functions.\\
            2. Determine the time intervals when the object is slowing down or speeding up.

            \vspace{0.5em}
            \textbf{Golden answer:}\\
            1. The velocity function $v(t)$ = $6\cdot t^2-6\cdot t-12$ and acceleration function $a(t)$ = $12\cdot t-6$ \\
            2. The time intervals when the object speeds up $\left(0,\frac{1}{2}\right)$, $(2,\infty)$ and slows down $\left(\frac{1}{2},2\right)$
        \examplesplit
                    \textbf{Reference solution:}\\

            The velocity is the derivative of the position function:
            
            \[
            v(t) = s'(t) = \boxed{6t^2 - 6t - 12}
            \]
            
            The acceleration is the derivative of the velocity function:
            
            \[
            a(t) = v'(t) = \boxed{12t - 6}
            \]
            
            To determine when the object is speeding up or slowing down, we compare the signs of \(v(t)\) and \(a(t)\). \\
            
            \textbf{\textit{Step 1: Find the Zeros of \(v(t)\) and \(a(t)\)}}
            
            First, solve for \(v(t) = 0\):
            
            \[
            6t^2 - 6t - 12 = 0 \quad \Rightarrow \quad t^2 - t - 2 = 0 \quad \Rightarrow \quad (t - 2)(t + 1) = 0
            \]
            
            Thus, \(t = 2\) and \(t = -1\). Next, solve for \(a(t) = 0\):
            
            \[
            12t - 6 = 0 \quad \Rightarrow \quad t = \frac{1}{2}
            \] \\

            \textbf{\textit{Step 2: Analyze the Signs of \(v(t)\) and \(a(t)\)}}
            
            We analyze the signs of \(v(t)\) and \(a(t)\) on the intervals determined by \(t = -1\), \(t = \frac{1}{2}\), and \(t = 2\).
            
            $$
            \begin{array}{cccc}
            \hline
            \text{Interval} & v(t) & a(t) & \text{Behavior} \\
            \hline
            (-\infty, -1) & >0 & <0 & \text{Slowing down} \\
            (-1, \frac{1}{2}) & <0 & <0 & \text{Speeding up} \\
            (\frac{1}{2}, 2) & <0 & >0 & \text{Slowing down} \\
            (2, \infty) & >0 & >0 & \text{Speeding up} \\
            \hline
            \end{array}
            $$ \\

            \textbf{\textit{Step 3: Account for non-negative time}}
            
            The object is speeding up on \( \boxed{\left(0, \frac{1}{2}\right) \text{ and } \left(2, \infty\right)} \) and slowing down on \( \boxed{\left(\frac{1}{2}, 2\right)} \).
        }
    \caption{A sample \umath problem, including the reference solution and the golden answer.}
    \label{fig:umath-sample}
\end{figure*}

\FloatBarrier\newpage
\subsection{\mumath Sample Problem} \label{appendix:meta-eval-tasks}

\begin{figure*}[!htb]
    \centering
    \examplebox{Example: Integral Calculus}{
        \textbf{\umath Problem:} \\
        Solve the integral:\\
        \[
            \int{\frac{20 \cdot \cos(-10 \cdot x)^3}{21 \cdot \sin(-10 \cdot x)^7} \, dx}
        \]

        \vspace{0.5em}
        \textbf{Golden answer:}\\
        \[
        C+\frac{1}{21}\cdot\left(\frac{1}{2}\cdot\left(\cot(10\cdot x)\right)^4+\frac{1}{3}\cdot\left(\cot(10\cdot x)\right)^6\right)
        \]
        
        \vspace{0.5em}
        \textbf{LLM-generated answer:}\\
        \[
            -\frac{3\sin(10x)^2 - 2}{126\sin(10x)^6} + C
        \]

        \examplesplit
        \vspace{0.5em}

        \textbf{Golden judge verdict:} Yes
        
        \vspace{1em}
        
        \textit{\textbf{Comment:}} \\
        \textit{Omitting the arbitrary constants, the reference and the submission could be expressed, respectively, as}
        \[
        \frac{\cot^6(10 x)}{63} + \frac{\cot^4(10 x)}{42} \ \ \
         \text{~\textit{and}~} \ \ \,
         \frac{\csc^6(10 x)}{63} - \frac{\csc^4(10 x)}{42},
        \]
        \textit{which differ by a constant term of} $1/126$.
    }
    \caption{A sample \mumath problem, illustrating the comparison between the golden and LLM-generated answers.}
    \label{fig:meta-eval-sample}
\end{figure*}






\FloatBarrier\newpage
\section{\umath Topic Distribution}

\umath covers a variety of topics across the six of its subjects. Table~\ref{table:umath_subtopics} presents the total number of topics per subject, along with the names and sample counts for the seven most populated topics in each.

\begin{table*}[h!]
    \centering
    \begin{adjustbox}{width=0.8\linewidth}
    \begin{tabular}{@{}lrl@{}}
    \toprule
    \textbf{Subject}             & \textbf{Sample Count} & \textbf{Topic} \\ \midrule
    \textbf{Differential Calculus} & 29 & Curve Sketching \\
    \textbf{(51 unique topics)}                               & 13 & Limits \\
                                   & 12 & One-Sided Limits \\
                                   & 12 & L'Hospital's Rule \\
                                   & 11 & Increasing and Decreasing Functions \\
                                   & 11 & Higher Derivatives \\
                                   & 10 & Applications of Derivatives (Local Extrema) \\\midrule
    \textbf{Sequences and Series}  & 40 & Taylor Series \\
    \textbf{(28 unique topics)}    & 30 & Fourier Series \\
                                   & 18 & Maclaurin Series \\
                                   & 12 & Approximating Constants Using Power Series \\
                                   &  6 & Radius of Convergence (Center of Convergence) \\
                                   &  5 & Differentiate Power Series \\
                                   &  4 & Error in Approximation \\ \midrule
    \textbf{Integral Calculus}     & 83 & The Substitution Rule \\
    \textbf{(35 unique topics)}    & 24 & Antiderivatives \\
                                   & 10 & Volumes of Solids of Revolution About the X-Axis \\
                                   &  9 & Trigonometric Substitutions and Inverse Substitutions \\
                                   &  9 & Integrate Respect Independent Variable \\
                                   &  7 & Applications of Integrals \\
                                   &  7 & Single Variable Surface Area Integrals \\ \midrule
    \textbf{Precalculus Review}    & 55 & Trigonometric Functions \\
    \textbf{(19 unique topics)}    & 24 & Zeros \\
                                   & 11 & Inverses of Functions \\
                                   &  8 & Inequalities \\
                                   &  7 & Equations with Exponents and Logarithms \\
                                   &  7 & Properties of Functions \\
                                   &  6 & Exponential Functions \\ \midrule
    \textbf{Algebra}               & 18 & Equations and Inequalities \\
    \textbf{(74 unique topics)}    & 13 & Polynomial Equations \\
                                   &  8 & Find Composition of Two Functions \\
                                   &  7 & Polynomials \\
                                   &  6 & Find Slope Line \\
                                   &  6 & Applications of Exponential Function \\
                                   &  6 & Quadratic Equations \\ \midrule
    \textbf{Multivariable Calculus}& 13 & Triple Integrals \\
    \textbf{(53 unique topics)}    & 11 & Lagrange Multipliers \\
                                   &  9 & Double Integrals in Polar Coordinates \\
                                   &  8 & Derivatives of Parametric Equations \\
                                   &  8 & Integrals of Multivariable Functions \\
                                   &  8 & Double Integral Over General Region \\
                                   &  6 & Classification of Critical Points \\
                                   \bottomrule
    \end{tabular}
    \end{adjustbox}
    \caption{Unique topic counts and top seven populated topics together with their sample sizes per subject.}
    \label{table:umath_subtopics}
\end{table*}

\FloatBarrier
\newpage
\section{Prompts}

\subsection{Prediction Prompt} \label{appendix:prediction}

\begin{figure*}[!htb]
    \centering
    
    \examplebox{Solution CoT Prompt}{
    \texttt{\{\{problem\_statement\}\}}\\
    Please reason step by step, and put your final answer within \symbol{92}boxed\{\}

    \examplesplit

    \vspace{0.5em}
    \textit{\textbf{Comment:}}\\
    \textit{Images, if present, are passed by way of a provider-native interface. \\
    For OpenAI-compatible endpoints this is done through the \texttt{image\_url} field.\footnote{\url{https://platform.openai.com/docs/guides/vision}}}
    }
    \caption{Inference prompt used for sampling solutions given the problem statements.}
\end{figure*}

    

    



\newpage
\subsection{Judgment Prompts} \label{appendix:judgment}

\begin{figure*}[!htb]
    \centering

    \examplebox{Judgment Automatic CoT Prompt}{
        You'll be provided with a math problem, a correct answer for it and a solution for evaluation. \\
        You have to answer whether the solution is correct or not. \\

        \texttt{-}\texttt{-}\texttt{-}\\
        PROBLEM STATEMENT:\\
        \texttt{\{\{problem\_statement\}\}} \\
        
        CORRECT ANSWER:\\
        \texttt{\{\{golden\_answer\}\}} \\
        
        SOLUTION TO EVALUATE:\\
        \texttt{\{\{model\_output\}\}}\\
        \texttt{-}\texttt{-}\texttt{-} \\
        
        Now please compare the answer obtained in the solution with the provided correct answer to evaluate whether the solution is correct or not. \\

        Think step-by-step, then conclude with your final verdict by putting either "Yes" or "No" on a separate line.
    }
    \caption{AutoCoT judgment prompt used for comparing sampled solutions to the golden labels. This prompt variant is only meant for \mumath experimentation and has not been used in \umath evaluation.} \label{appendix:judgement_prompt_autocot}
\end{figure*}

\begin{figure*}[!htb]
    \centering

    \examplebox{Judgment CoT Prompt}{
        You'll be provided with a math problem, a correct answer for it and a solution for evaluation. \\
        You have to answer whether the solution is correct or not. \\

        \texttt{-}\texttt{-}\texttt{-} \\
        PROBLEM STATEMENT: \\
        \texttt{\{\{problem\_statement\}\}} \\
        
        CORRECT ANSWER: \\
        \texttt{\{\{golden\_answer\}\}} \\
        
        SOLUTION TO EVALUATE: \\
        \texttt{\{\{model\_output\}\}} \\
        \texttt{-}\texttt{-}\texttt{-} \\

        Now please compare the answer obtained in the solution with the provided correct answer to evaluate whether the solution is correct or not.\\

        Think step-by-step, following these steps, don't skip any:\\
        1. Extract the answer from the provided solution\\
        2. Make any derivations or transformations that may be necessary to compare the provided correct answer with the extracted answer\\
        3. Perform the comparison\\
        4. Conclude with your final verdict --- put either "Yes" or "No" on a separate line
    }
    \caption{CoT judgment prompt used for comparing sampled solutions to the golden labels. This prompt variant is our default one, and also the one used for \umath evaluations.} \label{appendix:judgement_prompt_cot}
\end{figure*}

\newpage
\begin{figure*}[!htbp] 
    \centering

    \examplebox{Judgment Extract Prompt}{
        You'll be given a result of an evaluation of some mathematical solution by a professional evaluator. \\
        You need to extract the final verdict of this evaluation in simple terms: is the solution graded as correct or not. \\

        Output only a single label --- "Yes", "No" or "Inconclusive" --- according to the provided evaluation ("Yes" if the solution is graded as correct, "No" if the solution is graded as incorrect, "Inconclusive" if the evaluation is incomplete or the final verdict is not settled upon). \\

        Only output "Inconclusive" for incomplete or unsettled evaluations. If the evaluation does contain a single final verdict like "Yes", "Correct", "True", "No", "Incorrect", "False" and so on, even if it is supplied with some additional disclaimers and remarks, output a "Yes" or "No" label accordingly.\\
        
        Here goes your input:\\
        \texttt{`}\texttt{`}\texttt{`}\\
        \texttt{\{\{generated\_judgment\}\}}\\
        \texttt{`}\texttt{`}\texttt{`}\\
    
        Now please output exactly either "Yes", "No" or "Inconclusive".
    }
    \caption{Prompt for extracting the final verdict from the judge's output.} \label{appendix:judgement_eprompt}
\end{figure*}



\FloatBarrier\newpage
\section{\umath Visual Comparison}\label{appendix:umath_visual_comparison}

\begin{figure*}[!ht]
    \centering
    \includegraphics[width=1.0\textwidth]{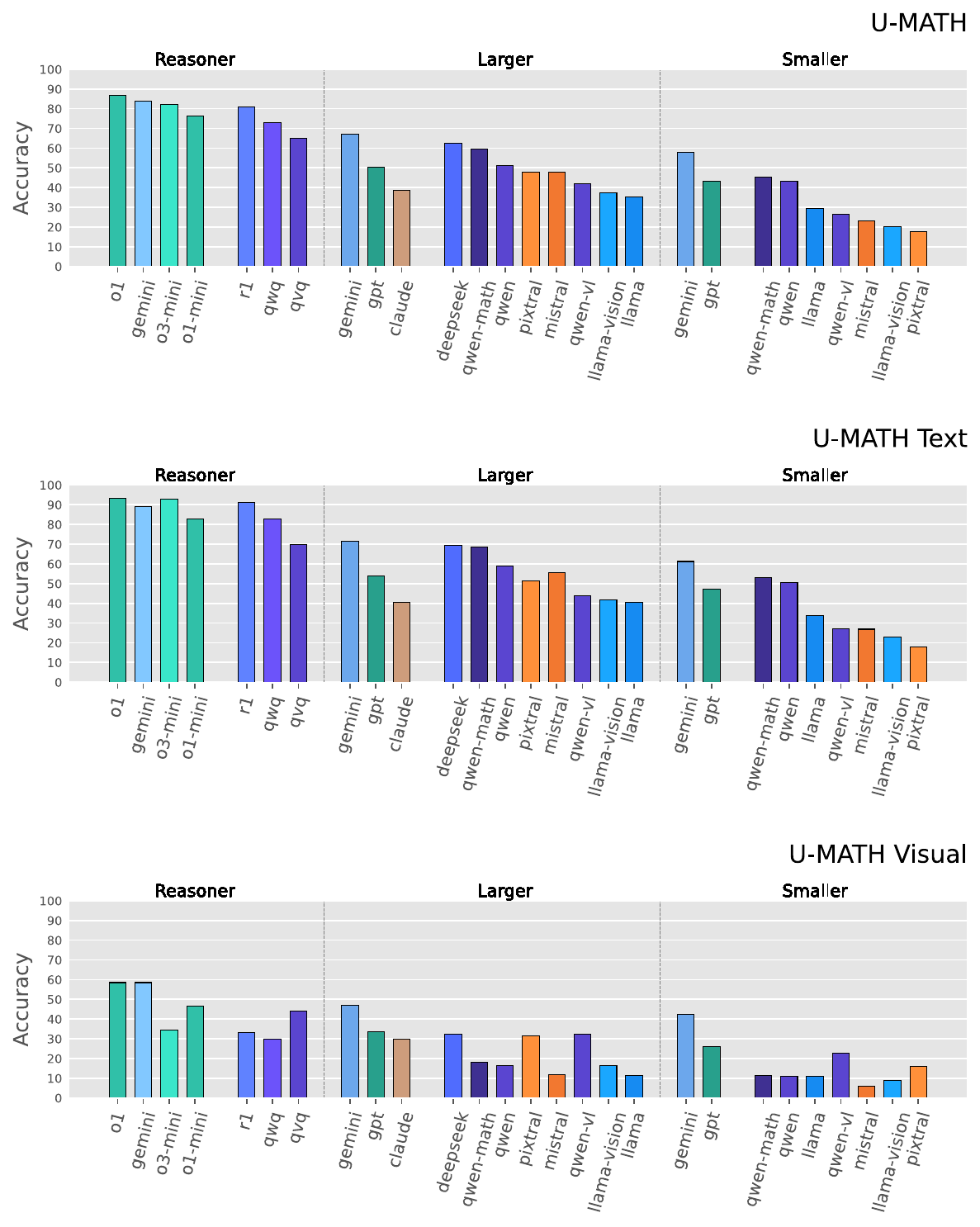}
    \caption{Performance of the selected top-performing models on \umath, \umathtext and \umathvisual.}
    \label{fig:models_chart}
\end{figure*}


    
    
    
    
    


\FloatBarrier\newpage
\section{\mumath Inconclusive Judgment Rate} \label{appendix:mumath_inconclusive}
\begin{table}[!ht]
    \centering
    \begin{adjustbox}{width=0.4\textwidth}
    \begin{tabular}{l||r|r}
    \toprule
    \textbf{Model} & \textbf{IncRate, AutoCoT} & \textbf{IncRate, CoT}   \\
    \midrule
    Llama-3.1 8B            & 13.4    & 22.9  \\
    Llama-3.1 70B           & 5.0     & 13.8  \\
    Ministral 8B              & 0.6     & 5.3   \\
    Mistral Large           & 0.4     & 1.7   \\
    Qwen2.5-Math 7B         & 2.8     & 2.4   \\
    Qwen2.5-Math 72B        & 1.2     & 0.7   \\
    Qwen2.5 7B              & 1.0     & 1.2   \\
    Qwen2.5 72B             & 1.6     & 2.1   \\
    DeepSeek-V3             & 0.2     & 0.2   \\
    GPT-4o-mini             & 0.0     & 0.1   \\
    GPT-4o                  & 0.0     & 0.0   \\
    Gemini 1.5 Flash        & 0.0     & 0.1   \\
    Gemini 1.5 Pro          & 0.0     & 0.0   \\
    Claude 3.5 Sonnet       & 0.0     & 0.0   \\
    QwQ-32B-Preview         & 0.6     & 0.9   \\
    Gemini 2.0 Flash Thinking & 0.2     & 0.5 \\
    DeepSeek-R1             & 0.0     & 0.3   \\
    o1-mini                 & 0.0     & 0.1   \\
    o1                      & 0.0     & 0.1   \\
    o3-mini                 & 0.0     & 0.0   \\
    \bottomrule
    \end{tabular}
    \end{adjustbox}
    \caption{Percentages of inconclusive judgments produced by each model under different prompting schemes on \mumath.}
    \label{table:mumath_inconlusive}
\end{table}

\FloatBarrier\newpage
\section{Problem-solving vs. Judgment, Conservatism vs. Leniency, Reasoning vs. Coherence}\label{appendix:mumath_vs_umath}

\begin{figure*}[!htbp]
    \centering
    \includegraphics[width=1.\textwidth]{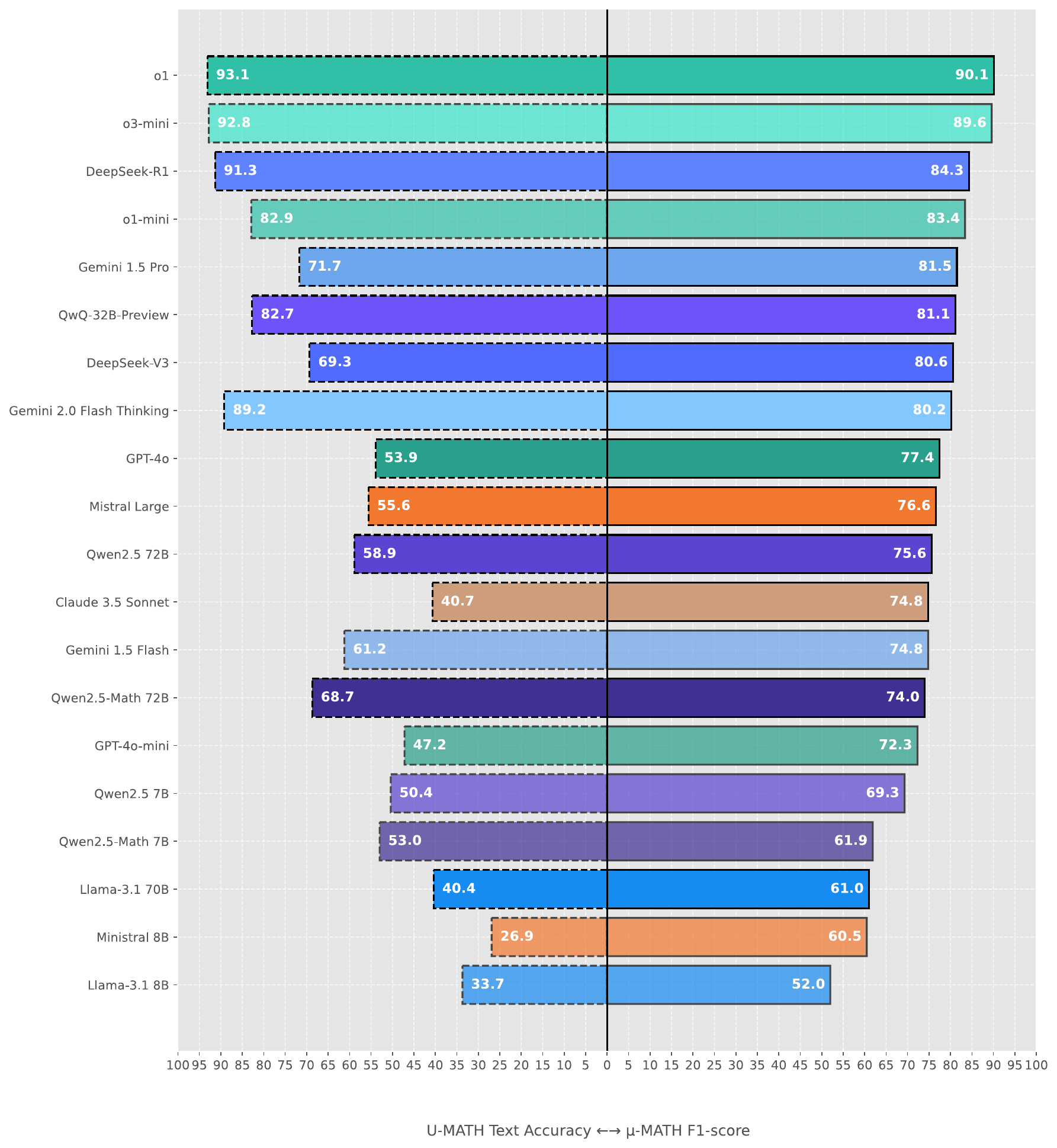}
    \caption{Comparison of LLMs' textual problem-solving (\umathtext) vs judgment (\mumath) performance.}
    \label{fig:mumath_vs_umath}
\end{figure*}

This section compares the performance of the models on \umathtext and \mumath. The overall score distribution shown in Figure~\ref{fig:mumath_vs_umath} reveals that improved problem-solving capabilities do not necessarily translate to improved judgment. Furthermore, the data suggest \textbf{a potential trade-off} between these capabilities, as observed with non-reasoning models, which exhibit a wedge-shaped trend: the two skills improve together up to a certain threshold, beyond which they appear inversely correlated.

Based on extensive manual examination, we propose this phenomenon reflects a trade-off between \textbf{formal domain-specific reasoning} and \textbf{general coherence}. This is perhaps best illustrated by considering the tradeoff's `extreme ends': Claude Sonnet achieves strong judgment scores despite significantly weaker problem-solving compared to models with similar judgment rankings, something allowing it to compensate for problem-solving deficit, while Qwen-Math, conversely, excels in problem-solving relative to neighbors, indicating some hindrance in translating problem-solving prowess into more effective judgment.


Studying the model responses suggests that what hinders Qwen-Math is exactly the inferior coherence: the model is generally struggling with instruction comprehension, adherence to formatting rules and `keeping track' of the tasks beyond mathematical problem-solving. Claude, by comparison, is excellent at all of those things, but often to the detriment of in-depth reasoning. To illustrate how this typically plays out, Appendix~\ref{appendix:mumath_demo} provides an example comparing the Claude's and Qwen's judgments on a single \mumath sample. Notice how Claude is restrictive and superficial in its comparison, whereas Qwen `loses the structure' along the way, designating only the first two steps prescribed with the CoT prompt (see prompt contents in Appendix~\ref{appendix:judgment}), omitting points three and four and switching to the `common problem-solving output style'.

\vspace{1em}
\noindent
We observe this dynamic with all the models to an extent, leading to two corresponding `judgment styles':
\begin{itemize}
  \item \textbf{Lenient judges:} tend to `follow the solution', are generally more verbose and good at going into involved derivation chains, which is necessary to arrive at a true positive verdict in more complex scenarios (higher TPR), but comes at a cost of increased hallucination risk and mislabeling negative examples (lower TNR).
  \item \textbf{Conservative judges:} tend to be more `anchored on the label', are generally more structured and precise, and also less heavy on long hallucination-prone outputs, which reduces the negative mislabeling (higher TNR) but comes at the expense of poor positive recall (lower TPR).
\end{itemize}

\vspace{1em}
Linking behavioral tendencies to typical outcomes allows us to quantify and visualize these patterns by decomposing the \mumath performance into TPR and TNR, as shown in Figure~\ref{fig:mumath_tpr-tnr}. Notice in particular that Claude and Qwen-Math appear as `the opposites` --- having respectively the highest overall TNR and highest overall TPR among the non-reasoners with an approximately equal F1-score. 

\vspace{1em}
\noindent
There are also other patterns emerging, offering deeper insight into the discussed trade-offs.
\begin{itemize}
    \item \textbf{Model tendencies run in the family}: for example, both of the GPT-4 models are conservative, as are both of the Gemini 1.5 models, while all the Qwen models tend to be more lenient. This suggests that these tendencies are largely induced by training data.
    \item \textbf{More balanced training leads to more balanced performance}, as evidenced by comparing the TPR-TNR ratio of Qwen2.5 and Qwen2.5-Math.
    \item \textbf{Losing in capability, on the contrary, exacerbates the bias}, with conservative models mainly losing in TPR and lenient models mainly losing in TNR when moving from a larger model to a smaller one (e.g. Gemini 1.5 Pro $\Rightarrow$ Gemini 1.5 Flash, GPT-4o $\Rightarrow$ GPT-4o-mini, Qwen2.5-72B $\Rightarrow$ Qwen2.5-7B). This indicates that not only a well-balanced training mixture is required but also adequate model capability to generalize over it.
    \item \textbf{Reasoner systems `push to the right'}, consistent with our observations that increased mathematical problem-solving and verbosity --- hallmarks of reasoner systems --- correlate with an increase in TPR\footnote{Notably, R1 is the only reasoning system that is closer to conservative models in terms of its scores. Upon inspection, we found that its reasoning traces are indeed often driving it towards conservative judgments, the model displaying `hyper-fixation' over minute details of the golden labels. This is the first case we encountered where an increase in coherence would probably aid more with true positives rather than true negatives. But the sentiment remains the same: coherence and reliability are required to appropriately and successfully apply problem-solving skills to the task at hand.}. This together with previous observations suggests that a potentially promising approach to improving the broad performance and coherence of reasoning systems lies in better training balancing. Notice also that the best performing models, o1 and o3-mini, are among the most balanced ones.
\end{itemize}






\FloatBarrier\newpage
\section{\mumath Judgment Examples} \label{appendix:mumath_demo}

\begin{figure*}[!htb]
\centering
\examplebox{Example: Differential calculus}{
    \textbf{\umath Problem:} \\
    Compute the derivative of the function
    $$y=\sqrt{\frac{ x^5 \cdot \left(2 \cdot x^6+3\right) }{ \sqrt[3]{1-2 \cdot x} }}$$

    \vspace{0.5em}
    \textbf{Golden answer:}\\
    \[
    y'=\frac{-128\cdot x^7+66\cdot x^6-84\cdot x+45}{-24\cdot x^8+12\cdot x^7-36\cdot x^2+18\cdot x}\cdot\sqrt{\frac{x^5\cdot\left(2\cdot x^6+3\right)}{\sqrt[3]{1-2\cdot x}}}
    \]
    
    \vspace{0.5em}
    \textbf{LLM-generated answer:}\\
    \[
    \sqrt{\frac{x^5 \cdot (2x^6 + 3)}{\sqrt[3]{1 - 2x}}} \cdot \frac{1}{2} \left( \frac{5}{x} + \frac{12x^5}{2x^6 + 3} + \frac{2}{3(1 - 2x)} \right)
    \]

    \examplesplit
    \vspace{0.5em}

    \textbf{Golden judge verdict:} Yes
}
\end{figure*}

\begin{figure*}[!htb]
\centering
\examplebox{Claude 3.5 Sonnet Judgment}{
1. The solution's answer is:
$\sqrt{\frac{x^5 \cdot (2x^6 + 3)}{\sqrt[3]{1 - 2x}}} \cdot \frac{1}{2} \left( \frac{5}{x} + \frac{12x^5}{2x^6 + 3} + \frac{2}{3(1 - 2x)} \right)$

2. The correct answer is:
$\frac{-128\cdot x^7+66\cdot x^6-84\cdot x+45}{-24\cdot x^8+12\cdot x^7-36\cdot x^2+18\cdot x}\cdot\sqrt{\frac{x^5\cdot\left(2\cdot x^6+3\right)}{\sqrt[3]{1-2\cdot x}}}$

3. To compare these expressions:
\begin{itemize}
\item Both have the same factor $\sqrt{\frac{x^5\cdot\left(2\cdot x^6+3\right)}{\sqrt[3]{1-2\cdot x}}}$
\item However, the other factors are completely different:
    \begin{itemize}
    \item Solution has: $\frac{1}{2} \left( \frac{5}{x} + \frac{12x^5}{2x^6 + 3} + \frac{2}{3(1 - 2x)} \right)$
    \item Correct answer has: $\frac{-128\cdot x^7+66\cdot x^6-84\cdot x+45}{-24\cdot x^8+12\cdot x^7-36\cdot x^2+18\cdot x}$
  \end{itemize}
\item These expressions are not algebraically equivalent, as they have different degrees in numerator and denominator
\end{itemize}
4. No
}
\end{figure*}

\newpage
\begin{figure*}[!htb]
\centering
\examplebox{Qwen2.5-Math 72B Judgment}{
    1. The solution provides the derivative as:
    \[
    \frac{dy}{dx} = \sqrt{\frac{x^5 \cdot (2x^6 + 3)}{\sqrt[3]{1 - 2x}}} \cdot \frac{1}{2} \left( \frac{5}{x} + \frac{12x^5}{2x^6 + 3} + \frac{2}{3(1 - 2x)} \right)
    \]
    
    2. The provided correct answer is:
    \[
    y' = \frac{-128x^7 + 66x^6 - 84x + 45}{-24x^8 + 12x^7 - 36x^2 + 18x} \cdot \sqrt{\frac{x^5 \cdot (2x^6 + 3)}{\sqrt[3]{1 - 2x}}}
    \]
    
    We need to simplify the expression inside the parentheses of the solution and see if it matches the fraction in the correct answer. First, let's rewrite the expression inside the parentheses:
    \[
    \frac{5}{x} + \frac{12x^5}{2x^6 + 3} + \frac{2}{3(1 - 2x)}
    \]
    
    To combine these fractions, we need a common denominator. The common denominator is
    \[
    x(2x^6 + 3)(3(1 - 2x)) = 3x(2x^6 + 3)(1 - 2x)
    \]
    Rewriting each term with this common denominator:
    \[
    \frac{5}{x} = \frac{5 \cdot 3(2x^6 + 3)(1 - 2x)}{3x(2x^6 + 3)(1 - 2x)} = \frac{15(2x^6 + 3)(1 - 2x)}{3x(2x^6 + 3)(1 - 2x)}
    \]
    \[
    \frac{12x^5}{2x^6 + 3} = \frac{12x^5 \cdot 3x(1 - 2x)}{3x(2x^6 + 3)(1 - 2x)} = \frac{36x^6(1 - 2x)}{3x(2x^6 + 3)(1 - 2x)}
    \]
    \[
    \frac{2}{3(1 - 2x)} = \frac{2 \cdot x(2x^6 + 3)}{3x(2x^6 + 3)(1 - 2x)} = \frac{2x(2x^6 + 3)}{3x(2x^6 + 3)(1 - 2x)}
    \]
    
    Combining these fractions:
    \[
    \frac{15(2x^6 + 3)(1 - 2x) + 36x^6(1 - 2x) + 2x(2x^6 + 3)}{3x(2x^6 + 3)(1 - 2x)}
    \]
    
    Expanding the numerator:
    \[
    15(2x^6 + 3)(1 - 2x) = 30x^6 - 60x^7 + 45 - 90x 
    \]
    \[
    2x(2x^6 + 3) = 4x^7 + 6x
    \]
    
    Adding these together:
    \[
    30x^6 - 60x^7 + 45 - 90x + 36x^6 - 72x^7 + 4x^7 + 6x = -128x^7 + 66x^6 - 84x + 45
    \]
    
    So the expression inside the parentheses simplifies to:
    \[
    \frac{-128x^7 + 66x^6 - 84x + 45}{3x(2x^6 + 3)(1 - 2x)}
    \]
    
    Therefore, the derivative is:
    \[
    \sqrt{\frac{x^5 \cdot (2x^6 + 3)}{\sqrt[3]{1 - 2x}}} \cdot \frac{1}{2} \cdot \frac{-128x^7 + 66x^6 - 84x + 45}{3x(2x^6 + 3)(1 - 2x)} = \frac{-128x^7 + 66x^6 - 84x + 45}{6x(2x^6 + 3)(1 - 2x)} \cdot \sqrt{\frac{x^5 \cdot (2x^6 + 3)}{\sqrt[3]{1 - 2x}}}
    \]
    
    Notice that:
    \[
    -24x^8 + 12x^7 - 36x^2 + 18x = -6x(4x^7 - 2x^6 + 6x - 3) = -6x(2x^6 + 3)(2x - 1) = 6x(2x^6 + 3)(1 - 2x)
    \]
    
    Thus, the two expressions are identical. Therefore, the solution is $\boxed{\textbf{correct}}$.
}
\label{fig:mumath_claude-vs-qwen}
\end{figure*}

\end{document}